\documentclass{article}

\usepackage[preprint]{neurips_2026}
\usepackage[utf8]{inputenc} % allow utf-8 input
\usepackage[T1]{fontenc}    % use 8-bit T1 fonts
\usepackage{hyperref}       % hyperlinks
\usepackage{url}            % simple URL typesetting
\usepackage{booktabs}       % professional-quality tables
\usepackage{amsfonts}       % blackboard math symbols
\usepackage{nicefrac}       % compact symbols for 1/2, etc.
\usepackage{microtype}      % microtypography
\usepackage{xcolor}         % colors
\usepackage{microtype}
\usepackage{graphicx}

\usepackage{algorithm}
\usepackage{algpseudocode}
\usepackage{amsmath}
\usepackage{multirow}
\title{When Integral Meets Decomposition: A Signal-Level Self-Supervised Feature Decompose Paradigm for Multi-Modal Image Fusion}

\author{%
  \textbf{Zeyu Wang}\textsuperscript{1}
  \qquad
  \textbf{Jiayu Wang}\textsuperscript{1, 2}
  \qquad
  \textbf{Haiyu Song}\textsuperscript{1, $^{*}$}
  \qquad
  \textbf{Haoran Duan}\textsuperscript{3, \thanks{Corresponding authors}}\\
  \textsuperscript{1}College of Computer Science and Engineering,
  Dalian Minzu University, Dalian, China\\
  \textsuperscript{2}School of Artificial Intelligence and Robotics,
  Hunan University, Hunan, China\\
  \qquad
  \textsuperscript{3} Department of Automation, Tsinghua University, Beijing, China\\
  \texttt{\{wangzeyu, shy\}@dlnu.edu.cn}
  \qquad
  \texttt{jiayuwang@hnu.edu.cn}
  \qquad
  \texttt{haoran.duan@ieee.org}\\
}

\begin{document}
\maketitle
\begin{abstract}

Multimodal image fusion (MMIF) aims to integrate complementary information from different modalities into a high-quality fused image and support downstream tasks. Recently, feature decomposition has become an important paradigm by separating source images into common and modality-specific unique features. However, existing methods lack clear supervision because ground-truth (GT) decomposition feature maps are unavailable. They usually combine multiple image-level metrics as losses, which are inherently incomplete and may conflict since each pixel couples attributes such as texture, edge, and contour. To address this, we propose a 1D signal-level self-supervised feature decomposition paradigm. Our core insight is to reformulate feature decomposition from unclear 2D image-level supervision into an integral-driven 1D signal-level optimization problem. This objective-level reformulation uses the 1D signal form to compute the integral constraint. The decomposer is optimized by the integral area between common and original signals, enabling more stable optimization with a clear optimization objective. Our model follows a two-stage SSL framework. Stage I designs dual pretext tasks for integral-driven decomposition at the signal level and structure-preserving reconstruction at the image level. Stage II fuses unique features and combines them with common features to reconstruct the fused image. Experiments on representative MMIF tasks show state-of-the-art (SOTA) performance. Code: github.com/Wangjiayu0512/SIDFusion.

\end{abstract}

\section{Introduction}

\begin{figure}
    \centering
    \setlength{\abovecaptionskip}{1pt}
    \includegraphics[width=1\linewidth]{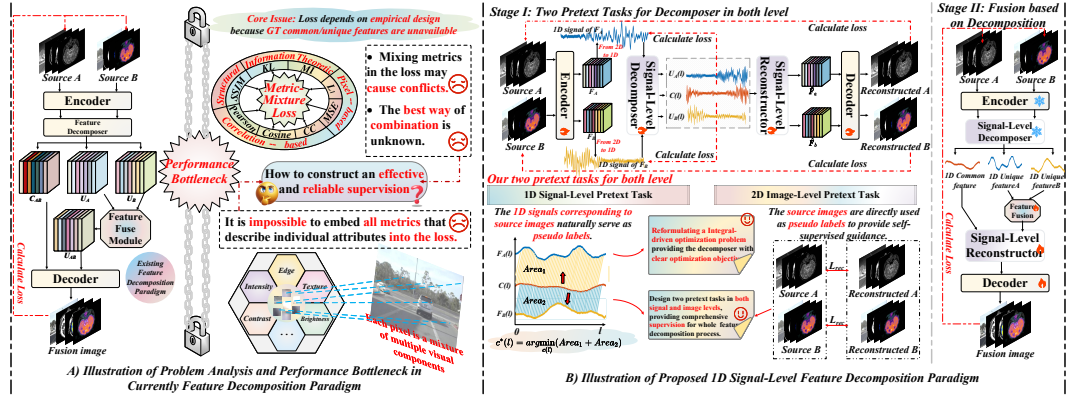}
    \caption{Existing decomposition paradigms \emph{vs} Ours. We introduce signal-level decomposition and image-level reconstruction pretext tasks for accurate decomposition and high-quality fusion.}
    \label{fig:pipline}
\end{figure}   

\label{sec:intro}
Multimodal image fusion (MMIF) integrates complementary information from different modalities to produce fused images~\cite{Tang2024DRMF, Tang2025ControlFusion, liu2024promptfusion, KARIM2023185}. Typical tasks include medical image fusion (MIF) and infrared–visible fusion (VIF) \cite{10812907, 10443302, 10874856, 10480582}, where modalities capture complementary cues, such as thermal radiation in infrared images and textures in visible images~\cite{li2025difiisr}. MMIF also benefits downstream tasks, including medical image segmentation~\cite{liu2022multimodal} and object detection~\cite{liu2025dcevo, liu2023bi}.

Currently, feature decomposition paradigm has become a key method~\cite{liang2022fusion}. It decomposes source images into two complementary feature types: common features representing shared scene information and unique features reflecting modality-specific details. This strategy facilitates the interaction and fusion of modality-specific information while preserving complete scene structure. 

However, existing 2D image-level methods face a key challenge that decomposer training heavily depends on handcrafted or empirical loss design caused by the lack of ground truth (GT) common/unique feature maps as supervision. Existing methods have to empirically embed multiple metrics into the loss function \cite{zhao2023cddfuse, bai2024tdfusion, bai2025refusion, cheng2025fdfuse, Deng2019deep, LI2026113022}. It is well known that each image pixel is a coupling of multiple visual attributes, such as edges, textures, contrast, light and contour. Obviously, it is almost impossible to incorporate all metrics that describe specific visual attributes into the loss, which makes an intrinsic difficulty. More importantly, we have no prior knowledge of whether these metrics are mutually conflicting, nor can we determine which combinations lead to better performance. This raises a central question: \textbf{How can we construct an effective and reliable supervisory objective?}

To address this, we propose a 1D signal-level self-supervised feature decomposition paradigm. Our core insight is to reformulate feature decomposition from an unclear supervision problem requiring simultaneous optimization of multiple visual attributes into an integral-driven mathematical problem. This reformulation shifts the focus from architectural design to the construction of a clear supervisory objective for feature decomposition. Accordingly, we design a tailored loss. In this formulation, one only needs to optimize the integral area between the common and source signals. It avoids conflicts from metric-mixing losses and enables disentanglement of coupled information. Under this integral-driven objective, the decomposer achieves more stable and accurate optimization.

Holistically, our model follows a two-stage self-supervised learning (SSL) framework \cite{simeoni2025dinov3}. In Stage I, we design dual pretext tasks at both the signal- and image-level. For decomposition, motivated by the reformulated problem, we design a novel signal-level pretext task. Features from source images are transformed into 1D signals and decomposed into low- and high-frequency components via DWT \cite{nechyba2004introduction}. The network then acts as an integral solver, estimating common features from paired low-frequency bands and extracting unique features from high-frequency bands of each modality. Paired source signals directly provide pseudo labels for strict mathematical constraints. For reconstruction, we return to the image-level, where original images supervise spatial structure preservation. In Stage II, we first fuse the unique feature signals, then combine them with the common signals, and finally apply IDWT to reconstruct fused images with complete information. Experimental results demonstrate that our method achieves state-of-the-art (SOTA) performance, validating the proposed paradigm.

Our contributions are summarized as follows:

$\bullet$ We propose a novel self-supervised feature decomposition paradigm at 1D signal-level to enable accurate feature decomposition.\\
$\bullet$ We reformulate the feature decomposition problem from an image-level supervision problem with unclear objectives into a signal-level integral optimization problem with an explicit mathematical objective. This clear optimization target reduces the ambiguity caused by empirical metric-mixture losses and enables more accurate feature decomposition.\\
$\bullet$ We design complementary dual pretext tasks at both levels. The signal-level task offers strong mathematical constraints for decomposition, while the image-level task ensures reliable spatial structures. They achieve both precise decomposition and faithful reconstruction.

\section{Related Work}

\label{sec:Related Work}
\textbf{Feature Decomposition-based MMIF.} Feature decomposition-based fusion methods decompose multimodal images into two complementary types of features: common features and unique features. This strategy not only promotes sufficient interaction and fusion of cross-modal complementary information, but also helps preserve the overall scene structure. Several methods adopt this idea. For instance, DIDFuse \cite{zhao2020didfuse} designs a dedicated feature decomposition module to explicitly separate common and unique features in an end-to-end manner; FD-Fuse \cite{cheng2025fdfuse} imposes indirect constraints on the decomposition process via pretext tasks; CU-Net \cite{Deng2019deep} formulates feature decomposition as a mathematical optimization problem and uses sparse convolutions to solve a hand-crafted objective function. Furthermore, external prior knowledge from large language models is used to guide the decomposition process. MTG-Fusion \cite{wang2025mtg} leverages a large language model to generate multiple descriptive texts and uses their embeddings to guide visual feature decomposition. However, since GT common and unique features are unavailable, these decomposers still lack reliable supervision. We address this issue by reformulating feature decomposition over signalized features as an integral-driven optimization problem, which provides an explicit objective for accurate feature separation.

\textbf{Self-Supervised Learning.} Self-supervised learning constructs pretext tasks to generate supervision from unlabeled data, usually through pretraining and downstream adaptation~\cite{10630605}. Due to the lack of GT in MMIF, many fusion methods adopt SSL strategies. DeFusion~\cite{liang2022fusion} learns cross-modal semantic relations via masked image restoration. SMFuse \cite{9369892} adopts a self-supervised scheme to generate accurate masks for multi-focus image fusion. Wang et al. use image super-resolution as a pretext task, treating high-resolution images as pseudo labels for low-resolution inputs \cite{wang2022self}. EMMA \cite{zhao2024equivariant} exploits image equivariance and generates supervision signals from geometrically transformed natural images. However, these methods are not specifically designed for the feature decomposition paradigm and cannot fundamentally resolve the absence of GT for common and unique features. 
Therefore, we design an SSL paradigm with dual pretext tasks. We transform the feature decomposition problem into an integral solving problem defined at the 1D signal-level, which naturally yields a pretext task with a clear mathematical objective, using 1D signals of the source images to provide direct supervision. At the image-level, we use the source images as supervision for a reconstruction task, thus providing comprehensive supervision for the entire feature decomposition process.

\textbf{Signal Modeling-based Image Fusion.}
Recently, SigFusion~\cite{Wang_Feng_Wang_Wang_Song_2026} proposes a unified self-supervised paradigm from the signal-level perspective, integrating dataset synthesis and image fusion into one framework. However, it is fundamentally different from our work. SigFusion uses signal-level modeling mainly to learn modality-specific signal distributions and synthesize large-scale training data, aiming to alleviate the lack of realistic training pairs. Our method uses signalization to reformulate the decomposition problem itself, focusing on the lack of direct supervision for common and unique features and providing the decomposer with an explicit integral-driven objective.
\section{Method}

This section describes the workflow of our proposed model. As shown in Fig. \ref{fig:pipline}, our model consists of an encoder $\zeta$, a signal-level feature decomposer $SD(\cdot)$, an MLP based feature fusion module $\delta$ and  a decoder $\xi$. Both the encoder and decoder are composed of Restormer blocks \cite{zhao2023cddfuse}. Following the self-supervised learning (SSL) paradigm, our model adopts a two-stage paradigm. Stage I performs signal-level decomposition and reconstruction pretraining to obtain accurate common and unique features. Stage II conducts cross-modal fusion on the decomposed unique features and combines them with the common features to generate the final fused image.
\begin{figure}
    \centering
    \includegraphics[width=1\linewidth]{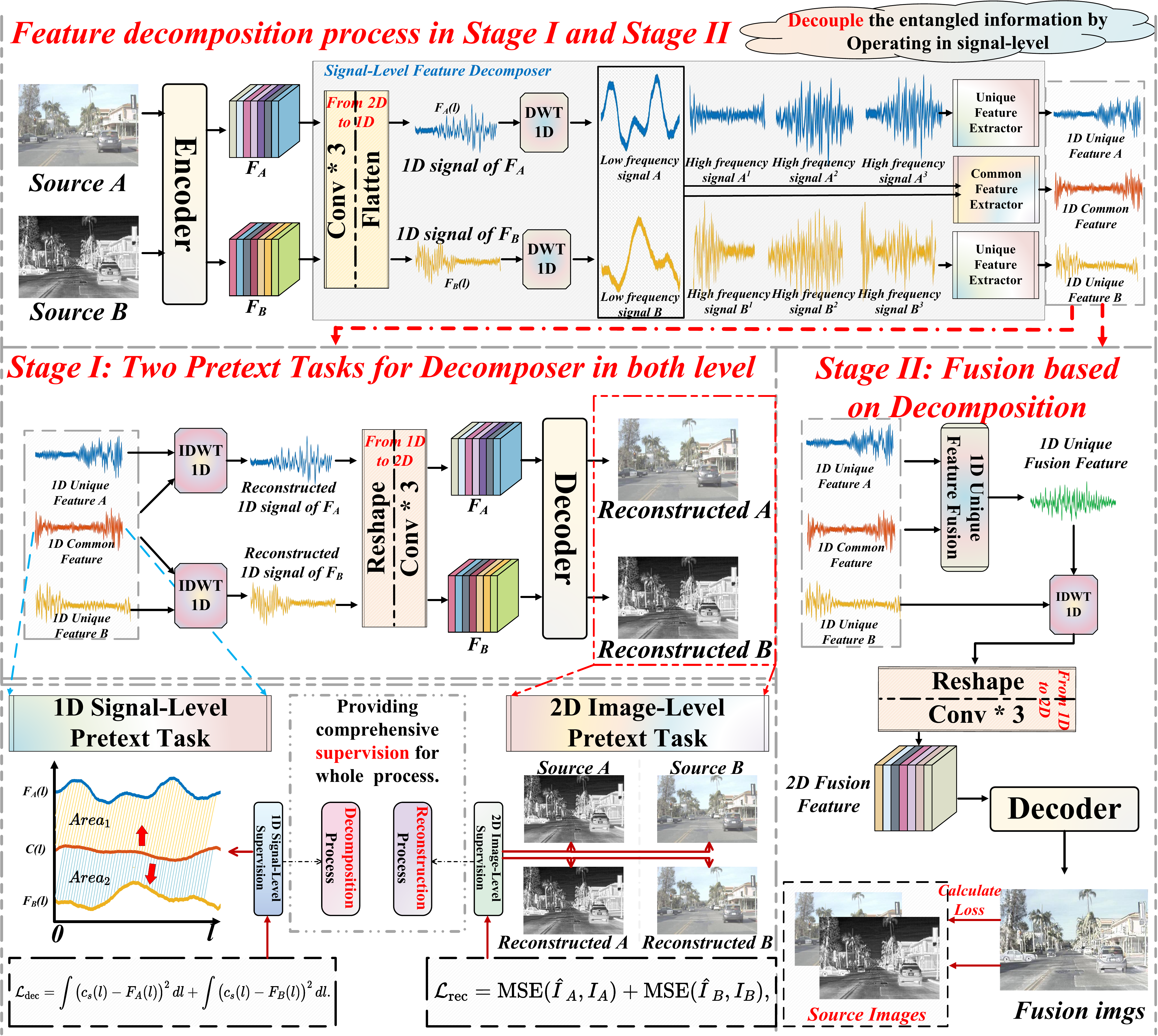}
    \caption{Schematic of the proposed model. It adopts a two-stage self-supervised paradigm. Stage I: feature decomposition via 1D signal-level and 2D image-level pretext tasks; Stage II: fuses the decomposed features to reconstruct the final fused image.}
    \label{fig:pipline}
\end{figure}
\subsection{Problem Formulation and Modeling}

In MMIF, mainstream methods based on feature decomposition paradigm typically follow a two-stage pipeline. Given a pair of source images $(I_A, I_B)$, a decomposer $D(\cdot)$ first extracts the common feature $C$ and unique features $(U_A, U_B)$:
\begin{equation}
C, U_A, U_B = D(I_A, I_B).
\end{equation}
Then, these features are fused to obtain fused image $I_f$:
\begin{equation}
I_f = F(C, U_A, U_B).
\end{equation}

However, these methods suffer from a fundamental limitation in supervision. Because ground-truth common and unique feature maps are unavailable, the training objective is inevitably constructed in a handcrafted or empirical manner, typically by mixing multiple attribute metrics. Yet each pixel couples diverse visual attributes such as edges, textures and contours, making it impractical to cover all attributes with metrics. Consequently, the final objective is often incomplete and may even introduce  conflicts, which prevents accurate and stable decomposition. Therefore, it is crucial to design a decomposition paradigm with a clear and reliable supervisory objective.

\textbf{Reformulating the Feature Decomposition Problem.}
To address these issues, we propose a signal-level feature decomposition paradigm that reformulates the decomposition problem from an ill-posed 2D image-level problem into a well-defined optimization problem in the 1D signal domain. 

Given a source image pair $(I_A, I_B)$ and their 1D signal representations $(s_A(l), s_B(l))$, we optimize the common signal $c_s(l)$ by minimizing the accumulated integral area induced by its deviation from the two original source signals:
\begin{equation}
c_s^{\ast}(l)
=
\arg\min_{c_s}
\int \big(c_s(l)-s_A(l)\big)^2 dl
+
\int \big(c_s(l)-s_B(l)\big)^2 dl .
\end{equation}

In this formulation, the common signal is encouraged to lie close to both source signals in the integral sense, so that it can capture the component consistently supported by the two modalities. Compared with image-level metric mixing, this objective provides a more direct criterion for common feature estimation and reduces the influence of local fluctuations or modality-specific disturbances.

\subsection{Stage I: Self-supervised Feature Decomposition with Dual Pretext Tasks}

In Stage I, we design two complementary pretext tasks for self-supervised common and unique feature extraction. The signal-level task provides an explicit objective for feature decomposition, while the image-level task preserves spatial structures through reconstruction. Procedure is as follows.

\textbf{Feature Decomposition based on the Signal-level Pretext Task.}\\
Given a pair of source images $I_A, I_B$, we first extract high-level feature maps via an encoder $\zeta$:
\begin{equation}
F_A, F_B = \zeta(I_A), \zeta(I_B),
\end{equation}
These features are then transformed into 1D signal-level representations:
\begin{equation}
s_A(l), s_B(l) = \mathcal{H}\big(F_A, F_B\big),
\end{equation}
where $\mathcal{H}$ denotes the transformation with a $1\times1$ convolution followed by flattening and $l$ indexes the 1D positions corresponding to spatial locations.
To enable more fine-grained feature decomposition, we decompose each signal into low- and high-frequency sub-bands via DWT 1D:
\begin{align}
L_m(l), \{H_m^i(l)\}_{i=1}^{N} &= \mathrm{DWT}\big(s_m(l)\big), \quad m \in \{A,B\}.
\end{align}
Low-frequency components mainly encode global structure shared across modalities, while high-frequency components capture unique details such as textures and edges.

Based on these characteristics, we design a signal decomposer $SD(\cdot)$ with two paths:
the common-path takes the low-frequency pair $(L_A(l), L_B(l))$ as input and estimates the common signal $c_s(l)$:
\begin{equation}
c_s(l)
= SD\big(L_A(l), L_B(l)\big).
\end{equation}

The unique-path takes high-frequency sub-bands of each modality to estimate unique signals $u_{s^A}(l)$ and $u_{s^B}(l)$:
\begin{equation}
u_{s^A}(l), u_{s^B}(l)
= SD\big(\{H_A^i(l)\}_{i=1}^N, \{H_B^i(l)\}_{i=1}^N\big).
\end{equation}

To optimize the common feature, we adopt a tailored integral-driven loss $\mathcal{L}_{\text{dec}}$:
\begin{equation}
\mathcal{L}_{\text{dec}}
= \int \big(c_s(l) - s_A(l)\big)^2 dl
+ \int \big(c_s(l) - s_B(l)\big)^2 dl,
\end{equation}
where the 1D signals $s_A(l)$ and $s_B(l)$ serve as pseudo labels. This loss enforces global consistency along the entire signal, suppressing local noise and guiding accurate estimation of $c_s(l)$.

\textbf{Image Reconstruction based on the Image-level Pretext Task.}\\
The reconstruction process mirrors the decomposition and uses the decomposed features to recover the original images, providing an image-level constraint.

First, $c_s(l)$ and $u_{s^m}(l)$ are fed into the inverse DWT, denoted as $\mathrm{DWT}^{T}$, to reconstruct the 1D signal:
\begin{align}
\hat{s}_m(l) = \mathrm{DWT}^{T}\big(c_s(l), u_{s^m}(l)\big), \quad m \in \{A,B\}.
\end{align}
Then, the reconstructed signals are mapped back to 2D feature maps by the inverse transformation $\mathcal{H}^{T}$ and reconstructed to the source images by a decoder $\xi$:
\begin{align}
\hat{I}_m = \xi\big(\mathcal{H}^{T}(\hat{s}_m(l))\big), \quad m \in \{A,B\}.
\end{align}
In this reconstruction process, we employ $I_A$ and $I_B$ act as pseudo labels to preserve spatial structure:
\begin{equation}
\mathcal{L}_{\text{rec}}
= \mathrm{MSE}(\hat{I}_A, I_A)
+ \mathrm{MSE}(\hat{I}_B, I_B),
\end{equation}

\textbf{Loss Function of Stage I.} The overall loss function of Stage I is defined as Eq. \ref{loss_stage1}:
\begin{equation}
\mathcal{L}_{\text{stage1}}
= \alpha \mathcal{L}_{\text{dec}}
+ \beta \mathcal{L}_{\text{rec}},
\label{loss_stage1}
\end{equation}
where $\alpha$ and $\beta$ are hyperparameters. Through these dual-pretext tasks, Stage I provides a clear optimization target for decomposition at 1D signal-level and enforces structural correctness in 2D image-level, producing accurate common and unique features for the subsequent fusion stage.

\subsection{Stage II: Feature Decomposition-based Image Fusion}

In Stage II, our goal is to generate a high-quality fused image based on the accurate decomposed features obtained from Stage I. Training and inference processes are as follows.\\
\textbf{Training.}
We freeze the encoder $\zeta$ and the signal decomposer $SD$ pretrained in Stage I. Given a source image pair $(I_A, I_B)$, we obtain the common signal $c_s(l)$ and unique signals $u_{s^A}(l)$ and $u_{s^B}(l)$ via the Stage I modules. We then fuse the unique features via an MLP-based feature fusion module $\delta$:
\begin{equation}
u_{\text{f}}(l) = \delta\big(u_{s^A}(l), u_{s^B}(l)\big),
\end{equation}

Next, the fused specific signal $u_{\text{f}}(l)$ and the common signal $c_s(l)$ are passed through the inverse DWT to obtain the fused 1D signal:
\begin{equation}
f_s(l) = \mathrm{DWT^T}\big(c_s(l), u_{\text{f}}(l)\big).
\end{equation}
This signal is mapped back to a 2D feature map by $\mathcal{H}^T$ and decoded to the fused image:
\begin{align}
I_f &= \xi(\mathcal{H}^T\big(f_s(l)\big)\big).
\end{align}

\textbf{Loss Function of Stage II. }Following mainstream practice, we use the source images as supervision to guide fusion. The Stage II loss is defined as:
\begin{equation}
\begin{aligned}
\mathcal{L}_{\text{stage2}}
= \gamma \mathcal{L}_{\text{SSIM}}(I_f; I_A, I_B)
+ \lambda \mathcal{L}_{L1}(I_f; I_A, I_B)
+ \omega \mathcal{L}_\text{Grad}(I_f; I_A, I_B),
\end{aligned}
\end{equation}
where $\mathcal{L}_{\text{SSIM}}$ is the SSIM-based structural similarity loss, $\mathcal{L}_{L1}$ is the spatial $L_1$ loss, $\mathcal{L}_\text{Grad}$ is the gradient-domain $L_1$ loss, and $\gamma$, $\lambda$, $\omega$ are hyperparameters.\\
\textbf{Inference.}
During inference, only the Stage II fusion pipeline is executed. Given a registered source image pair $(I_A, I_B)$, the model $\mathcal{G}$ directly produces the fused image $I_f$ in a single forward pass:
\begin{equation}
I_f = \mathcal{G}(I_A, I_B).
\end{equation}

\begin{figure*}
 \centering
  \includegraphics[width=0.99\textwidth]{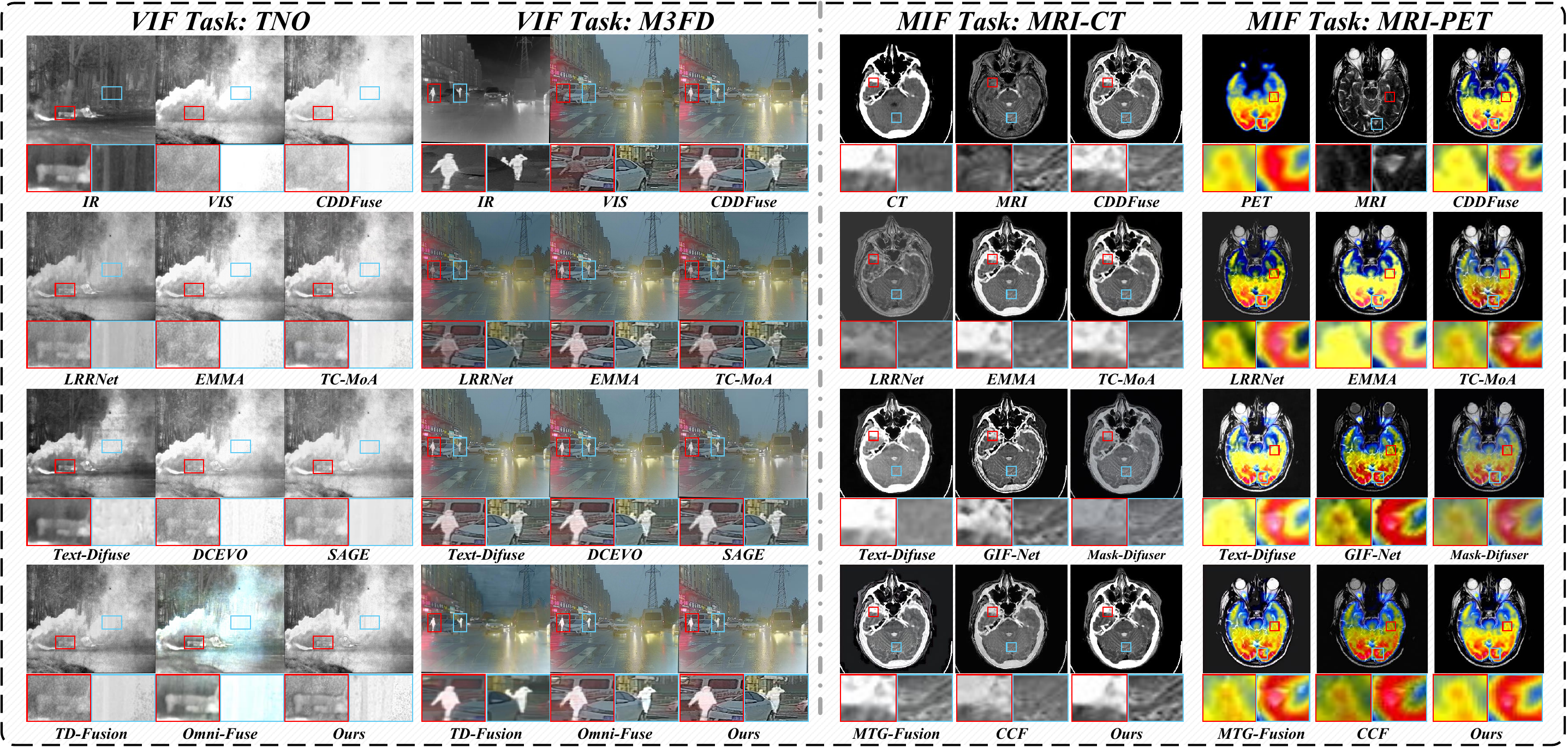}
  \caption{Qualitative comparison of various fusion models. }
  \label{Fig.compare}
\end{figure*}

\section{Experimental Results}
\subsection{Setting}
We adopt the AdamW optimizer with an initial learning rate of $3\times 10^{-4}$, batch size 4, and no weight decay. The learning rate is halved every 20 epochs. All experiments are conducted on Ubuntu 22.04 with an Intel i9-14900K CPU, RTX 4090 GPU, 64GB RAM, and PyTorch 2.3.0. In Stage I, we set $\alpha$ = 1, $\beta$ = 2 for 200 epochs. In Stage II, $\gamma$, $\lambda$ and $\omega$ are set as 1, 1 and 15.6 for 200 epochs, respectively.
\subsection{Datasets and Metrics}
\textbf{Datasets.}
For VIF, the training set consists of 400 image pairs from the MSRS\cite{Ma2022SwinFusion}. 
Testing is conducted on M$^3$FD (100 pairs), TNO (25 pairs), and MSRS (361 pairs).
For MIF, we construct the training set from 453 pairs of registered medical images from the Harvard Medical website \cite{Tang_2022_MATR}. 
For testing, we select 21 MRI–CT pairs \cite{zhao2023cddfuse}, 42 MRI–PET pairs \cite{zhao2023cddfuse}, and 73 MRI–SPECT pairs \cite{zhao2023cddfuse}. \\
\textbf{Metrics.}
We adopt widely used image quality metrics following mainstream practice, including $Q_{MI}$ \cite{qu2002information}, $Q_{NICE}$ \cite{wang2005nonlinear}, $Q_{P}$ \cite{xydeas2000objective}, $Q_{CB}$ \cite{wang2025highlight}, $MI$ \cite{qu2002information}, $VIF_P$ \cite{sheikh2006image}, and $Q_{Y}$ \cite{yang2008novel}. 
\subsection{Comparison with SOTA}
\textbf{Comparison Algorithms. }For VIF task, we select CDDFuse \cite{zhao2023cddfuse}, LRRNet \cite{li2023lrrnet}, EMMA \cite{zhao2024equivariant}, TC-MoA \cite{Zhu_2024_tcmoa}, Text-Difuse \cite{zhang2024textdif}, DCEvo \cite{liu2025dcevo}, SAGE \cite{wu2025SAGE}, TDFusion \cite{bai2024tdfusion}, Omni-Fuse \cite{zhang2025omnifuse}, C2RF \cite{Tang2024C2RF} and SigFusion\cite{Wang_Feng_Wang_Wang_Song_2026}. For MIF task, we select CDDFuse, LRRNet, EMMA, TC-MoA, Text-Difuse, CCF \cite{cao2024ccf},       BSA-Fusion \cite{li2025bsafusion}, Mask-Difuser \cite{Tang2024Mask-DiFuser}, MTG-Fusion \cite{wang2025mtg}, C2RF \cite{Tang2024C2RF} and SigFusion\cite{Wang_Feng_Wang_Wang_Song_2026}.\\
\textbf{Qualitative Comparison.}
Fig. \ref{Fig.compare} shows the qualitative comparison results of 12 fusion models on four test sets. More results are shown in Appendix \ref{sec:more compa}. For VIF task, the first 2 groups show results on TNO \cite{TOET2017249} and M$^3$FD \cite{liu2022target}. Benefiting from the 1D signal-level decomposition, our fused images better preserve both the common structure and modality-specific cues. For MIF task, the last two groups are the MRI-CT and MRI-PET datasets results. Our dual-pretext task design provides accurate and comprehensive supervision for both levels, yielding fused images with improved color fidelity.\\
\textbf{Quantitative Comparison.}
Table \ref{comparation_tab} reports the results of various models on seven metrics across VIF and MIF. Our model achieves superior results in most metrics. We attribute these gains to the signal-level decomposer and dual-pretext design, which provide clear supervision for the common component and enhance cross-modal interaction, yielding a high-quality representation for fusion.

\begin{table*}

\caption{Quantitative comparison of various fusion models on the VIF and MIF tasks. Best is \textbf{bold}; second best is \underline{underlined}. }
\centering
\resizebox{\textwidth}{!}{ 
\begin{tabular}{c|c|ccccccc|ccccccc|ccccccc}        
 \hline
\multicolumn{2}{c|}{\textbf{VIF Task}}  & \multicolumn{7}{c|}{\textbf{M$^3$FD}} &\multicolumn{7}{c|}{\textbf{MSRS}} &\multicolumn{7}{c}{\textbf{TNO}}  \\
 \hline
Methods & Pub/Year & $Q_{MI}$↑ & $Q_{NICE}$↑ &$ Q_P$↑ &$ Q_{CB}$↑ & $MI$↑ & $VIF_p$↑ & $Q_{Y}$↑ & $Q_{MI}$↑ & $Q_{NICE}$↑ &$ Q_P$↑ &$ Q_{CB}$↑ & $MI$↑ & $VIF_p$↑ & $Q_{Y}$↑& $Q_{MI}$↑ & $Q_{NICE}$↑ &$ Q_P$↑ &$ Q_{CB}$↑ & $MI$↑ & $VIF_p$↑ & $Q_{Y}$↑ \\
 \hline
CDDFuse & CVPR 23 & 0.5741 & 0.8123 & 0.4554 & 0.4746 & 3.8730 & 0.4410  & 0.8698 & 0.7576 & 0.8229 & 0.5399 & \underline{0.5669} & 4.9089 & 0.5106 & 0.8267 & 0.4654 & 0.8090 & 0.3789 & 0.4504 & 3.0691 & 0.4428 & 0.7874 \\
LRRNet & TPAMI 23 & 0.6353 & 0.8073 & 0.3709 & 0.4346 & 2.8124 & 0.3550  & 0.7209 & 0.6608 & 0.8082 & 0.3372 & 0.3928 & 2.8662 & 0.2933 & 0.5067 & 0.3686 & 0.8062 & 0.2303 & 0.4831 & 2.3868 & 0.3685 & 0.7011 \\
EMMA & CVPR 24 & 0.5564 & 0.8118 & 0.4602 & 0.4775 & 3.7712 & \underline{0.4690}  & 0.7979 & 0.6413 & 0.8161 & 0.4961 & 0.5432 & 4.1637 & \underline{0.5230} & 0.7840 & 0.4356 & 0.8081 & 0.3550 & 0.5112 & 2.9089 & 0.4847 & 0.8072 \\
TC-MoA & CVPR 24 & 0.5011 & 0.8096 & \underline{0.5133} & 0.4934 & 3.3330  & 0.4603 & 0.8538 & 0.7446 & 0.8119 & 0.4833 & 0.5558 & 3.4925 & 0.4924 & 0.8825 & 0.4068 & 0.8071 & 0.4020 & 0.5124 & 2.5747 & 0.4312 & \underline{0.8664} \\
Text-Difuse & NeurIPS 24 & 0.4262 & 0.8057 & 0.4323 & 0.3398 & 2.0535 & 0.1574 & 0.2566 & 0.7252 & 0.8032 & 0.5378 & 0.3660  & 1.3095 & 0.1570 & 0.2122 & 0.2977 & 0.8047 & 0.1167 & 0.4296 & 1.8773 & 0.2638 & 0.4410 \\

DCEvo & CVPR 25 & 0.6085 & 0.8135 & 0.4692 & 0.4801 & \underline{4.0074} & 0.4485 & 0.8987 & 0.6246 & 0.8170  & 0.4967 & 0.5627 & 4.0493 & 0.5107 & 0.8650  & \textbf{0.5739} & \underline{0.8125} & \underline{0.4280}  & 0.5089 & \underline{3.6054} & \underline{0.4605} & \textbf{0.9060}  \\
SAGE & CVPR 25 & 0.4566 & 0.8080  & 0.4338 & 0.4647 & 2.8508 & 0.4257 & 0.8324 & 0.5153 & 0.8101 & 0.4348 & 0.4775 & 2.9895 & 0.3455 & 0.7732 & 0.3751 & 0.8060  & 0.3323 & 0.4715 & 2.2033 & 0.3963 & 0.7855 \\
TD-Fusion & CVPR 25 & 0.4494 & 0.8081 & 0.4987 & \textbf{0.5108} & 3.0290  & 0.4629 & 0.8519 & 0.7462 & 0.8082 & 0.4170  & 0.5268 & 2.8434 & 0.4407 & 0.7922 & 0.3640  & 0.8062 & 0.3512 & \underline{0.5288} & 2.4507 & 0.4317 & 0.8183 \\
Omni-fuse & TPAMI 25 & 0.4882 & 0.8092 & 0.2770  & 0.4776 & 3.2075 & 0.4432 & 0.6543 & 0.7208 & 0.8079 & 0.5174 & 0.4211 & 2.6641 & 0.4121 & 0.4927 & 0.3814  & 0.8062  & 0.1908  & 0.4374  & 2.2871 & 0.3933 & 0.6406 \\
C2RF & IJCV 25 & 0.4082  & 0.8075  & 0.4989  & 0.3882  & 2.4979  & 0.2241  & 0.4293  & 0.6928  & 0.8039  & 0.4922  & 0.4246  & 1.7292  & 0.1589  & 0.3892  & 0.3535  & 0.8059  & 0.1847  & 0.4362  & 2.0637  & 0.2975  & 0.6155  \\
SigFusion & AAAI 26 & \underline{0.7123} & \underline{0.8186} & 0.4665 & 0.4796 & \textbf{4.5751} & 0.4478 & \textbf{0.9239} & \underline{0.8034} & \underline{0.8256} & \textbf{0.5492} & 0.5643 & \underline{4.9123} & 0.4432  & \underline{0.8772}  & 0.5296 & 0.8114 & 0.3936 & 0.5064  & 3.2423 & 0.4455 & 0.8911 \\
 \hline
Ours & - & \textbf{0.6608}  & \textbf{0.8209}  & \textbf{0.5735}  & \underline{0.4998}  & 4.1050  & \textbf{0.4694}  & \underline{0.8988}  & \textbf{0.8405}  & \textbf{0.8290}  & \underline{0.5483}  & \textbf{0.5798}  & \textbf{5.4547}  & \textbf{0.5253}  & \textbf{0.8936}  & \underline{0.5711}  & \textbf{0.8128}  & \textbf{0.4318}  & \textbf{0.5346}  & \textbf{3.8015}  & \textbf{0.4768}  & 0.8450  \\
 \hline
\multicolumn{2}{c|}{\textbf{MIF Task}}& \multicolumn{7}{c|}{\textbf{MRI-CT}} &\multicolumn{7}{c|}{\textbf{MRI-PET}} &\multicolumn{7}{c}{\textbf{MRI-SPECT}}  \\
 \hline
Methods & Pub/Year & $Q_{MI}$↑ & $Q_{NICE}$↑ &$ Q_P$↑ &$ Q_{CB}$↑ & $MI$↑ & $VIF_p$↑ & $Q_{Y}$↑ & $Q_{MI}$↑ & $Q_{NICE}$↑ &$ Q_P$↑ &$ Q_{CB}$↑ & $MI$↑ & $VIF_p$↑ & $Q_{Y}$↑& $Q_{MI}$↑ & $Q_{NICE}$↑ &$ Q_P$↑ &$ Q_{CB}$↑ & $MI$↑ & $VIF_p$↑ & $Q_{Y}$↑ \\
 \hline
CDDFuse & CVPR 23 & \underline{0.8403}  & \underline{0.8098}  & 0.4122  & \underline{0.6911}  & \underline{3.7522}  & 0.3735  & \underline{0.9047}  & 0.7856  & 0.8064  & 0.4448  & \textbf{0.7213}  & 2.7359  & 0.3872  & \textbf{0.9125}  & \underline{0.8778}  & 0.8068  & \underline{0.6289}  & \textbf{0.7305}  & 2.7296  & 0.4480  & \underline{0.9118}  \\
LRRNet & TPAMI 23 & 0.6273  & 0.8066  & 0.2073  & 0.2205  & 2.7570  & 0.2295  & 0.3285  & 0.6427  & 0.8051  & 0.3809  & 0.2129  & \textbf{3.6627}  & 0.2807  & 0.3122  & 0.7355  & 0.8056  & 0.3580  & 0.1839  & \textbf{3.8263}  & 0.2607  & 0.2763  \\
EMMA & CVPR 24 & 0.7032  & 0.8078  & 0.3160  & 0.5627  & 3.2790  & 0.3575  & 0.7590  & 0.6648  & 0.8055  & 0.3177  & 0.5606  & 2.4481  & 0.3577  & 0.7433  & 0.7086  & 0.8057  & 0.3368  & 0.4834  & 2.3611  & 0.3660  & 0.6633  \\
TC-MoA & CVPR 24 & 0.7279  & 0.8080  & 0.3686  & 0.5838  & 3.3169  & 0.3751  & 0.7782  & 0.6758  & 0.8055  & 0.4692  & 0.6129  & 2.4498  & 0.3900  & \underline{0.8394}  & 0.7397  & 0.8059  & 0.5564  & 0.5426  & 2.3331  & 0.4422  & 0.7659  \\
Text-Difuse & NeurIPS 24 & 0.6456  & 0.8075  & 0.2643  & 0.3239  & 3.1594  & 0.3150  & 0.3891  & 0.5235  & 0.8051  & 0.3118  & 0.2606  & 2.3310  & 0.2973  & 0.2630  & 0.5638  & 0.8053  & 0.3649  & 0.1976  & 2.2253  & 0.3261  & 0.2550  \\
CCF & NeurIPS 24 & 0.7361  & 0.8079  & 0.2825  & 0.3479  & 3.2844  & 0.2968  & 0.4329  & 0.6439  & 0.8050  & 0.2733  & 0.2637  & 2.4926  & 0.2854  & 0.3114  & 0.7520  & 0.8058  & 0.3911  & 0.2372  & 2.9410  & 0.3356  & 0.3177  \\

BSAFusion & AAAI 25 & 0.6929  & 0.8079  & 0.3229  & 0.3905  & 3.2802  & 0.3435  & 0.4578  & 0.7470  & \underline{0.8068}  & \underline{0.4788}  & 0.3446  & 2.8615  & 0.4197  & 0.4200  & 0.8150  & \underline{0.8072}  & \textbf{0.6393}  & 0.3106  & 2.7563  & \textbf{0.4550}  & 0.3766  \\
Mask-Difuser & TPAMI 25 & 0.7171  & 0.8081  & 0.3595  & 0.3443  & 3.2990  & 0.3088  & 0.4206  & 0.6911  & 0.8062  & 0.4399  & 0.3035  & 2.7060  & 0.3676  & 0.3590  & 0.7334  & 0.8062  & 0.5022  & 0.2620  & 2.5270  & 0.3677  & 0.3300  \\
MTG-fusion & IJCV 25 & 0.6610  & 0.8076  & 0.3896  & 0.3466  & 3.1782  & 0.2873  & 0.4343  & 0.5958  & 0.8053  & 0.3977  & 0.2651  & 2.3778  & 0.3043  & 0.3453  & 0.6129  & 0.8053  & 0.3987  & 0.2673  & 2.3096  & 0.2844  & 0.3084  \\
C2RF & IJCV 25 & 0.6374  & 0.8072  & 0.1306  & 0.3464  & 2.9530  & 0.2188  & 0.3220  & 0.5827  & 0.8050  & 0.2842  & 0.3543  & 2.1825  & 0.2205  & 0.2686  & 0.6760  & 0.8055  & 0.5087  & 0.3513  & 2.2566  & 0.3184  & 0.2887  \\
SigFusion & AAAI 26 & 0.8176  & 0.8086  & \textbf{0.6689}  & 0.3560  & 3.4913  & \underline{0.3819}  & 0.9063  & \underline{0.8762}  & 0.8074  & 0.4965  & \underline{0.7157}  & 3.0495  & \textbf{0.4610}  & 0.4273  & 0.8310  & \textbf{0.8083}  & 0.5162  & 0.6621  & 2.8941  & 0.4418  & 0.8798  \\
\hline
Ours & - & \textbf{0.9369}  & \textbf{0.8108}  & 0.4034  & \textbf{0.7008}  & \textbf{4.0184}  & \textbf{0.4034}  & \textbf{0.9439}  & \textbf{0.8843}  & \textbf{0.8074}  & \textbf{0.4967}  & 0.4740  & \underline{3.0547}  & \underline{0.4577}  & 0.5808  & \textbf{0.9778}  & \textbf{0.8083}  & 0.5507  & \underline{0.6697}  & \underline{3.0655}  &\underline{0.4520}  & \textbf{0.9149}  \\
\hline
\end{tabular}
}
\label{comparation_tab}
\end{table*}

\begin{table*}[t]
\centering
\caption{Quantitative results of ablation experiments on VIF and MIF tasks. }
\label{tab:ablation_two_blocks}
\resizebox{\textwidth}{!}{%
\begin{tabular}{c|c|c|ccccccc | c|c|c|ccccccc}
\toprule
\multicolumn{10}{c|}{\textbf{VIF Task}} & \multicolumn{10}{c}{\textbf{MIF Task}} \\
\cmidrule(lr){1-10} \cmidrule(lr){11-20}
Task & Case & Configurations & $Q_{MI}\!\uparrow$ & $Q_{NICE}\!\uparrow$ & $Q_{P}\!\uparrow$ & $Q_{CB}\!\uparrow$ & $MI\!\uparrow$ & $VIF_p\!\uparrow$ & $Q_{Y}\!\uparrow$
& Task & Case & Configurations & $Q_{MI}\!\uparrow$ & $Q_{NICE}\!\uparrow$ & $Q_{P}\!\uparrow$ & $Q_{CB}\!\uparrow$ & $MI\!\uparrow$ & $VIF_p\!\uparrow$ & $Q_{Y}\!\uparrow$ \\
\midrule

\multirow{9}{*}{VIF}
& \multirow{5}{*}{I}
& w/o signal-level Dec. & 0.6282 & 0.8181 & 0.5350 & 0.4716 & 3.8631 & 0.4391 & 0.8729
& \multirow{9}{*}{MIF}
& \multirow{5}{*}{I}
& w/o signal-level Dec. & 0.9012 & 0.8071 & 0.3627 & 0.6715 & 3.6129 & 0.3689 & 0.9127 \\
& & w/o DWT & 0.6425 & 0.8192 & 0.5551 & 0.4860 & 3.9812 & 0.4534 & 0.8847
& & & w/o DWT & 0.9185 & 0.8090 & 0.3825 & 0.6880 & 3.8310 & 0.3897 & 0.9312 \\
& & w/ IDWT $\leftarrow$ learnable Rec. & \underline{0.6531} & \underline{0.8198} & \underline{0.5631} & \underline{0.4925} & \underline{4.0200} & \underline{0.4594} & \underline{0.8911}
& & & w/ IDWT $\leftarrow$ learnable Rec. & \underline{0.9253} & \underline{0.8096} & \underline{0.3928} & \underline{0.6905} & \underline{3.9025} & \underline{0.3954} & \underline{0.9361} \\
& & w/ signal- $\leftarrow$ image-level Dec. & 0.6380 & 0.8185 & 0.5460 & 0.4793 & 3.9022 & 0.4468 & 0.8781
& & & w/ signal- $\leftarrow$ image-level Dec. & 0.9108 & 0.8086 & 0.3755 & 0.6810 & 3.7420 & 0.3798 & 0.9275 \\
& & Default setups (ours) & \textbf{0.6608} & \textbf{0.8209} & \textbf{0.5735} & \textbf{0.4998} & \textbf{4.1050} & \textbf{0.4694} & \textbf{0.8988}
& & & Default setups (ours) & \textbf{0.9369} & \textbf{0.8108} & \textbf{0.4034} & \textbf{0.7008} & \textbf{4.0184} & \textbf{0.4034} & \textbf{0.9439} \\
\cmidrule(lr){2-10} \cmidrule(lr){12-20}

& \multirow{4}{*}{II}
& w/o signal-level pretext task & 0.6315 & 0.8188 & 0.5420 & \underline{0.4868} & 3.9033 & 0.4512 & 0.8799
& & \multirow{4}{*}{II}
& w/o signal-level pretext task & 0.9144 & 0.8093 & 0.3786 & \underline{0.6869} & 3.7562 & 0.3831 & 0.9293 \\
& & w/o image-level pretext task & \underline{0.6495} & \underline{0.8195} & \underline{0.5525} & 0.4820 & \underline{3.9800} & \underline{0.4553} & \underline{0.8850}
& & & w/o image-level pretext task & \underline{0.9228} & \textbf{0.8294} & \underline{0.3866} & 0.6765 & \underline{3.8041} & \underline{0.3912} & \underline{0.9335} \\
& & w/ $L_{\mathrm{dec}} \leftarrow L_{1}$ & 0.6463 & 0.8195 & 0.5482 & 0.4817 & 3.9521 & 0.4520 & 0.8832
& & & w/ $L_{\mathrm{dec}} \leftarrow L_{1}$ & 0.9202 & 0.8091 & 0.3814 & 0.6851 & 3.7820 & 0.3886 & 0.9318 \\
& & Default setups (ours) & \textbf{0.6608} & \textbf{0.8209} & \textbf{0.5735} & \textbf{0.4998} & \textbf{4.1050} & \textbf{0.4694} & \textbf{0.8988}
& & & Default setups (ours) & \textbf{0.9369} & \underline{0.8108} & \textbf{0.4034} & \textbf{0.7008} & \textbf{4.0184} & \textbf{0.4034} & \textbf{0.9439} \\
\bottomrule
\end{tabular}%
}
\end{table*}

\subsection{Ablation Study}
To comprehensively verify the effectiveness of each key component, we conduct a systematic ablation study. More results are in Appendix \ref{sec:more abl}.\\
\textbf{Effect of Signal-Level Decomposer. }
Our signal-level decomposer transfers the decomposition problem into 1D signal-level. To assess its impact, we conduct the following ablations:
(1) removing the signal-level decomposer(Dec.); (2) removing frequency decomposition; (3) replacing IDWT with a learnable reconstructor(Rec.); (4) substituting an image-level decomposer; (5) default setup. The results are presented in Table \ref{tab:ablation_two_blocks} Case I, which demonstrate the effectiveness of our paradigm.

\textbf{Effect of Dual Pretext Tasks. }
To evaluate the dual pretext tasks at both levels, we perform the following ablations: (1) removing the signal-level pretext task;
(2) removing the image-level pretext task;
(3) replacing the proposed $L_{dec}$ with vanilla $L_1$ only;
(4) default configuration.
The results are shown in Table \ref{tab:ablation_two_blocks} Case II. Apparently, removing either pretext task degrades performance, highlighting the necessity to provide decomposer a clear supervision.

\textbf{Sensitivity to Key Hyperparameters.}
We analyze the sensitivity of key hyperparameters. Full analysis and ablation results are provided in Appendix \ref{sensitivity}.

\begin{figure}
 \centering
  \includegraphics[width=1\textwidth]{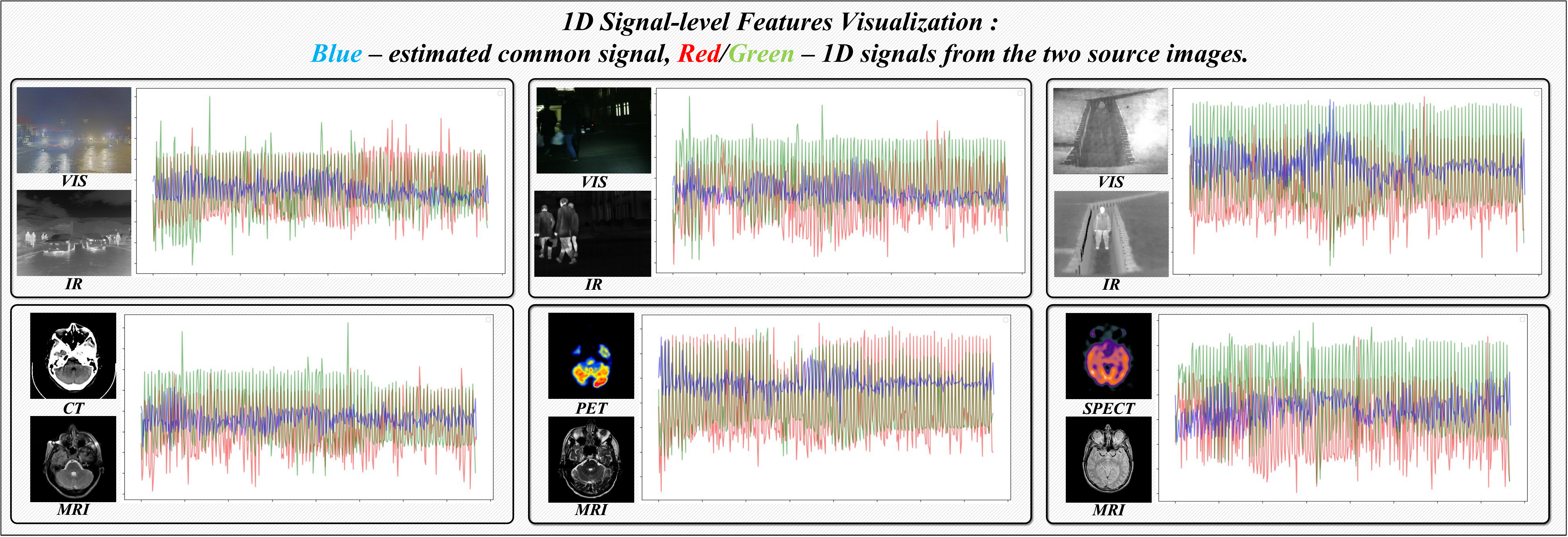}

\caption{Visualization of intermediate 1D signals. The learned common signal $c_s(l)$ follows the shared trend of the source signals $s_A(l)$ and $s_B(l)$.}

  \label{fig:signal_vis}
\end{figure}

\begin{figure*}
    \centering
    \includegraphics[width=1\linewidth]{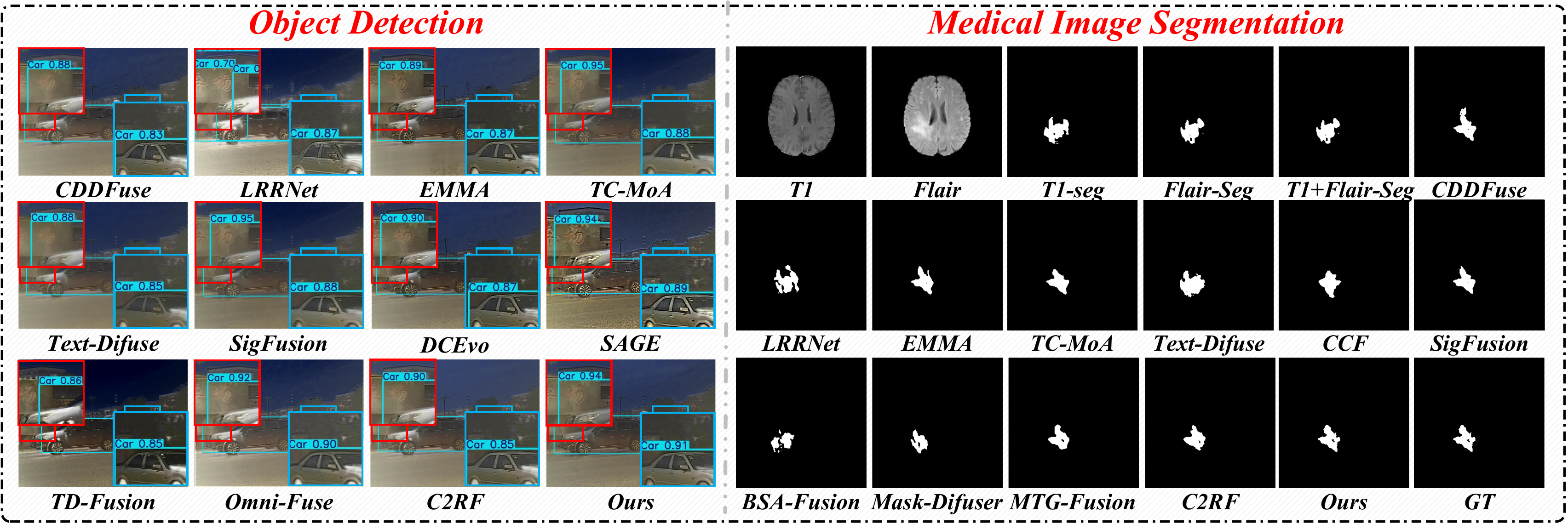}
    \caption{Qualitative results on downstream tasks. Left: object detection performance using fused images generated by different fusion methods. Right: medical image segmentation. Our fused images provide more accurate results.}
    \label{fig:downstream}
\end{figure*}

\subsection{Signal-Level Feature Visualizations}

To further validate the proposed signal-level formulation, we visualize the source signals $s_A(l)$ and $s_B(l)$ and the learned common signal $c_s(l)$ in Fig.~\ref{fig:signal_vis}. The red and green curves denote the two source signals, while the blue curve denotes the estimated common signal. It can be observed that $c_s(l)$ generally lies between $s_A(l)$ and $s_B(l)$ and follows their shared trend, indicating that it captures the component consistently shared by both modalities.

\subsection{Downstream Tasks}
\textbf{Object Detection for VIF Task. }
We evaluate detection on fused images produced by various VIF models. A pre-trained detector (YOLOv12) \cite{tian2025yolov12} is used to ensure fairness. Qualitative results are given in Fig.~\ref{fig:downstream} (left) and quantitative results see Appendix Table~\ref{tab:det_ap}. Compared with other fusion methods, our model achieves stronger detection performance, suggesting that it better preserves task-relevant complementary cues and provides more favorable inputs for downstream object detection.

\textbf{Medical Image Segmentation for MIF Task. }
We evaluate medical image segmentation using UniverSeg \cite{butoi2023universeg}. Qualitative results are given in Fig.~\ref{fig:downstream} (right) and quantitative results see Appendix Table~\ref{tab:seg_neworder}.
Compared with single-modality inputs, fused images provide more informative representations for segmentation, leading to predictions closer to the ground truth. Our method also outperforms other fusion models, indicating that the proposed decomposition strategy helps preserve common anatomical or lesion structures while retaining modality-specific details for boundary delineation.

% requires \usepackage{xcolor}
\iffalse
\begin{figure}
 \centering
  \includegraphics[width=1\textwidth]{PDFs/broader.pdf}

\caption{Broader impact of our paradigm. Integrating our signal-level decomposition into existing decomposition-based methods yields refined variants (denoted by “*”) that obtain fused images with clearer structures and more salient target details, as highlighted by the red and yellow circles.}

  \label{Fig.broader-1}
\end{figure}
\fi

\subsection{Broader Impact}

\textbf{Refining decomposition-based MMIF paradigms.}
Our signal-level decomposition paradigm provides a clearer objective for common and unique feature separation and can be plugged into existing decomposition-based MMIF models. To validate this, we insert it into DIDFuse, CDDFuse, FD-Fuse~\cite{cheng2025fdfuse}, and C2RF without changing their default settings. As shown in Fig.~\ref{Fig.broader-1}, the enhanced models preserve more complete structures and richer details. Quantitative results in Appendix Table~\ref{tab:broader_plug_in} further confirm its effectiveness in improving existing decomposition-based MMIF methods.

\begin{figure}
 \centering
  \includegraphics[width=1\textwidth]{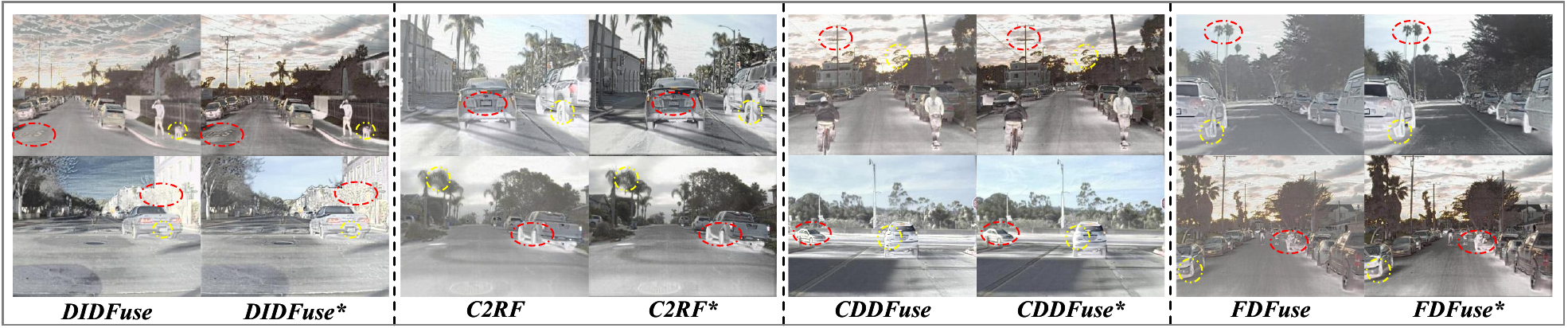}

\caption{Broader impact of our paradigm. Refined variants (denoted by “*”) obtain fused images with clearer structures and more salient target details, as highlighted by the red and yellow circles.}

  \label{Fig.broader-1}
\end{figure}

\begin{figure}
 \centering
  \includegraphics[width=1\textwidth]{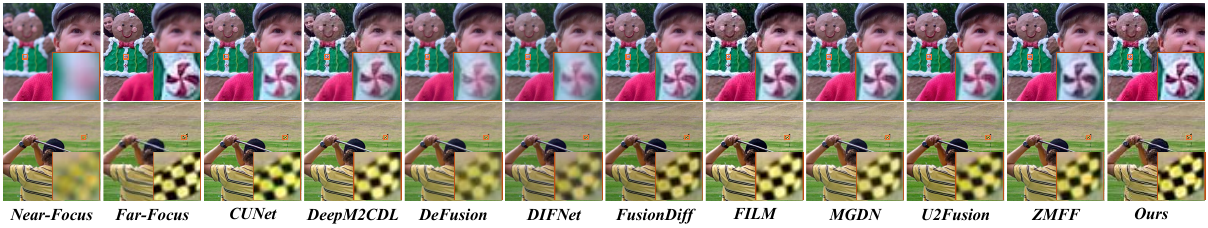}

\caption{Qualitative results on multi-focus image fusion, showing that the proposed signal-level decomposition paradigm can be naturally transferred to other image fusion tasks.}

  \label{Fig.broader-2}
\end{figure}

\textbf{Extension to other image fusion tasks.}
Our paradigm is not limited to MMIF. We further select multi-focus image fusion as a representative task, where common scene structures and source-specific focused details need to be separated. As shown in Fig.~\ref{Fig.broader-2}, the proposed paradigm can also produce clearer focus boundaries and more faithful structural details on this task. Detailed quantitative results are provided in the Appendix Table~\ref{tab:mfif}, suggesting its applicability to broader image fusion problems.

\subsection{Model Complexity and Overhead Analysis}

We analyze the computational overhead including parameters, GFLOPs, and inference time. Results are in Appendix Table~\ref{tab:complexity}. Our model remains computationally competitive. Although the DWT operation introduces a slight additional overhead, the overall complexity is still acceptable.

\section{Conclusion}
In this paper, we presented a signal-level feature decomposition paradigm that reformulates feature decomposition into an integral-driven optimization at the 1D signal-level. This reformulation yields a matched loss, providing a clear learning objective. Based on this, our two-stage SSL pipeline first performs a decomposition and reconstruction in Stage I, with dual pretext tasks at both levels supervising the entire decomposition and reconstruction process. Stage II then carries out fusion. Extensive experiments show advanced performance, validating the effectiveness of our paradigm.

\section*{Acknowledgments}
This work was supported by the National Natural Science Foundation of China [No.62401097, 62601484]; Fundamental Research Funds for Central Universities, Dalian Minzu University [No.0854-53]; Liaoning Province Applied Basic Research Program [2026JH2/101300163]; Liaoning Province Science and Technology Joint Plan (2024JH2/102600113).
% \bibliography{example_paper}
{
    \small
    % \normalem
    \bibliographystyle{abbrv}
    \bibliography{example_paper.bib} 
}
%%%%%%%%%%%%%%%%%%%%%%%%%%%%%%%%%%%%%%%%%%%%%%%%%%%%%%%%%%%%
\newpage
\appendix

\section{Motivation of the Signal-level Formulation}

The core motivation of adopting the signal-level formulation is to redefine the optimization objective for feature decomposition. For decomposition-based MMIF methods, GT feature maps for common and unique features are unavailable. Existing methods usually rely on a combination of image-level metrics, such as intensity and gradient, to indirectly constrain feature decomposition. However, these metrics only describe partial image attributes and are difficult to directly guide the decomposer to explicitly learn common feature and unique features. Therefore, the feature decomposition process still suffers from an unclear supervision target.

We reformulate this supervision-ambiguous feature decomposition problem as a signal-level integral optimization problem. The integral objective describes the global accumulated relation along the signal sequence, making the 1D signal space a natural domain for this optimization form. In this domain, the common signal can be constrained by the integral area between itself and the two source signals, which provides the decomposer with an explicit, computable, and decomposition-oriented learning target. In this way, the signalized representation becomes a key form to support the integral-driven objective, giving the decomposer a clearer optimization objective.

Please note that, the signal-level formulation does not discard spatial information. The 1D signal is mainly used to construct a clearer decomposition objective, while the decomposed signals are mapped back to 2D feature maps and further constrained by the image-level reconstruction task. Therefore, the signal-level and image-level tasks are complementary: the former provides an explicit objective for common and unique feature separation, while the latter preserves spatial structural consistency during reconstruction.

\section{Ambiguous Supervision of Pixel-Level Metric-Mixture Paradigm. }

In current decomposition-based MMIF methods, ground-truth common and unique feature maps are unavailable. 
As a result, existing methods usually constrain the decomposition process by imposing losses between the learned representation and the source images. 
However, each image pixel does not correspond to a single visual property. 
Instead, it simultaneously couples multiple attributes, such as intensity, contrast, edge, texture, contour, structure, and salient modality-dependent responses. 
To make this issue more explicit, we summarize representative visual attributes coupled in image pixels and their commonly used constraints in Table~\ref{tab:pixel_coupled_attributes}.

\begin{table*}[h]
\centering
\small
\caption{
Representative visual attributes coupled in image pixels and their commonly used constraints in MMIF. 
}
\label{tab:pixel_coupled_attributes}
\resizebox{\textwidth}{!}{
\begin{tabular}{l p{5.2cm} p{6.8cm}}
\toprule
Visual Attribute 
& Pixel-level Manifestation 
& Commonly Used Constraints \\
\midrule

Intensity / Brightness 
& Absolute pixel magnitude, local luminance 
& $L_1$ loss, $L_2$ loss, reconstruction loss \\

Contrast 
& Relative intensity difference between a pixel and its local neighborhood, or between foreground and background 
& SSIM contrast term, local contrast loss, entropy-based or information-based constraints \\

Edge 
& Sharp local intensity variation around object boundaries and structural transitions 
& Gradient loss, Sobel loss, Laplacian loss\\

Texture 
& fine details, and high-frequency responses 
& Gradient loss, frequency-domain loss \\

Boundary 
& Continuous object outlines and region boundaries 
& SSIM loss, gradient loss, boundary-aware loss \\

Structural Layout 
& Spatial arrangement of objects, organs, roads, targets, or anatomical regions 
& SSIM loss, perceptual loss \\

Color
& Channel-wise appearance, pseudo-color distribution, and modality-related color responses 
& Color consistency loss, histogram loss \\

\bottomrule
\end{tabular}
}
\end{table*}

As shown in Table~\ref{tab:pixel_coupled_attributes}, a single pixel simultaneously reflects multiple visual attributes, whereas each loss term only constrains part of them from a specific perspective. 
Obviously, it is almost impossible to embed all these attribute-related constraints into a single loss formulation. 
More importantly, there is no prior knowledge about which combination of these metrics is optimal for feature decomposition. 
Different losses may even impose potentially conflicting optimization preferences. 
Therefore, existing supervision based on empirical metric mixtures is inherently incomplete and ambiguous. 
This is exactly why we reformulate feature decomposition into a signal-level integral optimization problem, which provides the decomposer with a clearer objective.

% \hfill$\square$
\section{Two-Stage Training and Inference Pipeline}
For clarity, we summarize the overall pipeline of the proposed method in a two-stage manner, as shown in Algorithm~\ref{alg:stage1} and Algorithm~\ref{alg:stage2}.
\begin{algorithm}[t]
\caption{Stage I: Self-supervised signal-level feature decomposition with dual pretext tasks}
\label{alg:stage1}
\textbf{Input:} Training set $\mathcal{D} = \{(I_A^k, I_B^k)\}_{k=1}^N$ \\
\textbf{Modules:} Encoder $\zeta$, signal-level decomposer $SD$, decoder $\xi$ \\
\textbf{Hyper-parameters:} $\alpha, \beta$ for Stage I.
\begin{algorithmic}[1]
\State Initialize parameters of $\zeta$, $SD$, $\xi$
\For{epoch $= 1$ to $E_1$}
  \For{each mini-batch $(I_A, I_B) \subset \mathcal{D}$}
    % \STATE // \emph{High-level feature extraction}
    \State $F_A, F_B = \zeta(I_A), \zeta(I_B)$

    \State $s_A(l), s_B(l) = H(F_A, F_B)$ 
    
    \State $L_A(l), \{H_A^i(l)\}_{i=1}^N = \mathrm{DWT}(s_A(l))$
    \State $L_B(l), \{H_B^i(l)\}_{i=1}^N = \mathrm{DWT}(s_B(l))$
    \State // \emph{Signal-level decomposer: common \& unique signals}
    \State $c_s(l) = SD_{\text{common}}(L_A(l), L_B(l))$
    \State $u_s^A(l), u_s^B(l) = SD_{\text{specific}}(\{H_A^i(l)\}, \{H_B^i(l)\})$
    \State // \emph{Signal-level pretext: common-signal optimization with $L_{\mathrm{dec}}$}
    \State $L_{\mathrm{dec}} = \text{IntegralSquaredError}(c_s(l), s_A(l), s_B(l))$
    \State // \emph{Image-level pretext: reconstruct each source image}
    \State $\hat{s}_A(l) = \mathrm{DWT}^{-1}(c_s(l), u_s^A(l))$
    \State $\hat{s}_B(l) = \mathrm{DWT}^{-1}(c_s(l), u_s^B(l))$
    \State $\hat{F}_A = H^{\top}(\hat{s}_A(l)), \quad \hat{F}_B = H^{\top}(\hat{s}_B(l))$
    \State $\hat{I}_A = \xi(\hat{F}_A), \quad \hat{I}_B = \xi(\hat{F}_B)$
    \State $L_{\mathrm{rec}} = \mathrm{MSE}(\hat{I}_A, I_A) + \mathrm{MSE}(\hat{I}_B, I_B)$
    \State // \emph{Dual-pretext loss for decomposer}
    \State $L_{\text{stage1}} = \alpha L_{\mathrm{dec}} + \beta L_{\mathrm{rec}}$
    \State Update parameters of $\zeta$, $SD$, $\xi$ by back-propagating $L_{\text{stage1}}$
  \EndFor
\EndFor
\State \textbf{Output:} Pretrained encoder $\zeta$ and signal-level decomposer $SD$
\end{algorithmic}
\end{algorithm}
\subsection{Stage I: Self-supervised signal-level decomposition.}
In the first stage (Algorithm~\ref{alg:stage1}), we focus on learning a stable signal-level decomposer in a fully self-supervised fashion.
Given paired source images $(I_A, I_B)$, we first extract high-level features via the encoder $\zeta$, and then project the 2D feature maps into 1D signals.
These 1D signals are decomposed by a 1D DWT into low- and high-frequency components, which are further processed by the signal-level decomposer $SD$ to obtain a common signal $c_s(l)$ and modality-specific unique signals $u_s^A(l)$ and $u_s^B(l)$.
To train $SD$, we introduce a dual-pretext objective: a signal-level loss $L_{\mathrm{dec}}$ that directly supervises the decomposition in the 1D domain, and an image-level reconstruction loss $L_{\mathrm{rec}}$ that reconstructs each source image via inverse DWT and the decoder $\xi$.
The decomposer and encoder are optimized jointly with the combined loss $L_{\text{stage1}} = \alpha L_{\mathrm{dec}} + \beta L_{\mathrm{rec}}$, yielding a pretrained encoder–decomposer pair that can reliably separate common and modality-specific information.

\subsection{Stage II: Fusion based on signal-level decomposition.}
In the second stage (Algorithm~\ref{alg:stage2}), we freeze the encoder $\zeta$ and decomposer $SD$ learned in Stage~I, and only train the fusion head $\delta$ together with the decoder $\xi$.
For each training pair, we reuse the signal-level decomposition to obtain the common signal $c_s(l)$ and unique signals $u_sA(l)$ and $u_sB(l)$, and then feed the unique signals into the learnable fusion module $\delta$ to produce a fused unique signal $u_f(l)$.
Combining $c_s(l)$ and $u_f(l)$ through inverse DWT gives a fused 1D signal, which is mapped back to 2D feature space and decoded into the fused image $I_f$.
The fusion-specific loss $\mathcal{L}$ is computed between $I_f$ and the source images, encouraging the fused output to preserve structural details, intensity information, and gradient cues from both modalities.
After training, inference reduces to a single forward pass through this Stage~II pipeline, i.e., $I_f = G(I_A, I_B)$, for both MIF and VIF tasks.
\begin{algorithm}[t]
\caption{Stage II: Fusion based on signal-level decomposition}
\label{alg:stage2}
\begin{algorithmic}[1]
\Require Training set $\mathcal{D} = \{(I_A^k, I_B^k)\}_{k=1}^N$ // \emph{registered multimodal image pairs}
\Require Pretrained encoder $\zeta$ and signal-level decomposer $SD$ from Stage~I // \emph{fixed feature extractor and decomposer}
\Ensure Fused image generator $G(I_A, I_B)$ // \emph{for both MIF and VIF tasks}

\State Freeze parameters of $\zeta$ and $SD$ // \emph{use Stage~I as a fixed decomposer}
\State Initialize fusion module $\delta$ and decoder $\xi$ // \emph{learnable fusion head and image reconstructor}

\For{each training epoch}
  \For{each mini-batch $(I_A, I_B) \subset \mathcal{D}$}
    \State $F_A, F_B = \zeta(I_A), \zeta(I_B)$ 
    \State $s_A(l), s_B(l) = H(F_A, F_B)$ 
    \State $L_A(l), \{H_A^i(l)\} = \mathrm{DWT}(s_A(l))$ 
    \State $L_B(l), \{H_B^i(l)\} = \mathrm{DWT}(s_B(l))$ 
    \State $c_s(l) = SD_{\mathrm{common}}(L_A(l), L_B(l))$ 
    \State $u_s^A(l), u_s^B(l) = SD_{\mathrm{specific}}(\{H_A^i(l)\}, \{H_B^i(l)\})$ // \emph{modality-specific unique signals}
    \State $u_f(l) = \delta(u_s^A(l), u_s^B(l))$ // \emph{fuse unique signals by the learnable fusion head}
    \State $f_s(l) = \mathrm{DWT}^{-1}(c_s(l), u_f(l))$ // \emph{reconstruct fused 1D signal with inverse DWT}
    \State $F_f = H^{\top}(f_s(l))$ 
    \State $I_f = \xi(F_f)$ // \emph{decode fused feature into fused image}
    \State $L_{\mathrm{fuse}} = \mathcal{L}(I_f; I_A, I_B)$ 
    \State Update parameters of $\delta$ and $\xi$ using gradient of $L_{\mathrm{fuse}}$ // \emph{optimize fusion head and decoder}
  \EndFor
\EndFor

\State $I_f = G(I_A, I_B)$ for a given test pair $(I_A, I_B)$ // \emph{inference: one forward pass through the Stage~II pipeline}

\end{algorithmic}
\end{algorithm}
\begin{figure*}
    \centering
    \includegraphics[width=1\linewidth]{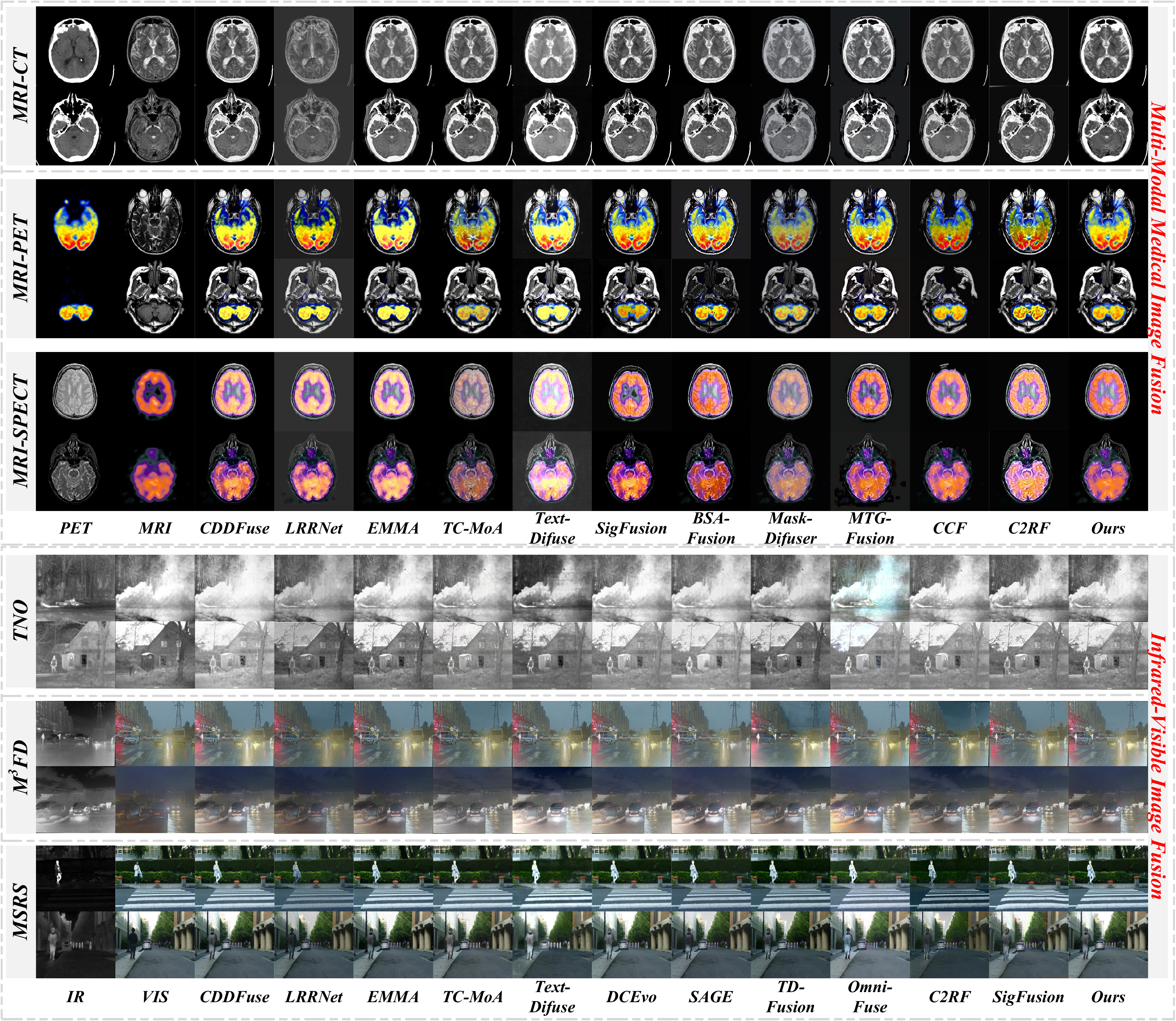}
    \caption{Full qualitative comparisons on six datasets. 
The top three groups show MIF task and the bottom three groups show VIF cases. In each row, the first two columns are the source image pairs and the remaining columns are fused results of different methods, with our method in the rightmost column, yielding sharper structures and richer complementary details.} 
    \label{fig:compare}
\end{figure*}
\section{Extended Experimental Results and Visualizations}
We adopt the same evaluation as in the main paper. 
Following mainstream practice, we use a set of widely used image quality metrics, including $Q_{MI}$~\cite{qu2002information}, $Q_{NICE}$~\cite{wang2005nonlinear}, $Q_{P}$~\cite{xydeas2000objective}, $Q_{CB}$~\cite{wang2025highlight}, $MI$~\cite{qu2002information}, $VIF_P$~\cite{sheikh2006image}, and $Q_{Y}$~\cite{yang2008novel}. 
For fair comparison, we also keep the same state-of-the-art baselines as in the main paper: for the VIF task we consider CDDFuse~\cite{zhao2023cddfuse}, LRRNet~\cite{li2023lrrnet}, EMMA~\cite{zhao2024equivariant}, TC-MoA~\cite{Zhu_2024_tcmoa}, Text-Difuse~\cite{zhang2024textdif}, SigFusion~\cite{Wang_Feng_Wang_Wang_Song_2026}, DCEvo~\cite{liu2025dcevo}, SAGE~\cite{wu2025SAGE}, TDFusion~\cite{bai2024tdfusion}, Omni-Fuse~\cite{zhang2025omnifuse}, and C2RF~\cite{Tang2024C2RF}; for the MIF task we adopt CDDFuse~\cite{zhao2023cddfuse}, LRRNet~\cite{li2023lrrnet}, EMMA~\cite{zhao2024equivariant}, TC-MoA~\cite{Zhu_2024_tcmoa}, Text-Difuse~\cite{zhang2024textdif}, CCF~\cite{cao2024ccf}, SigFusion~\cite{Wang_Feng_Wang_Wang_Song_2026}, BSA-Fusion~\cite{li2025bsafusion}, Mask-Difuser~\cite{Tang2024Mask-DiFuser}, MTG-Fusion~\cite{wang2025mtg}, and C2RF~\cite{Tang2024C2RF}. 
Due to page constraints in the main paper, only a subset of results can be shown there; this section reports extended quantitative results and richer visualizations across all datasets and modalities to more comprehensively demonstrate the behavior and advantages of our method.

\subsection{Full Qualitative Comparisons}
\label{sec:more compa}
We provide visual comparisons against all competing methods on six datasets.
For each dataset, we show the source images, the fused results generated by representative competitors, and the output of our method.
The extended visualizations are presented in Fig.~\ref{fig:compare}, where the first three groups correspond to MIF cases (MRI-CT, MRI-PET, MRI-SPECT) and the last three groups correspond to VIF cases (TNO, M$^3$FD, MSRS).
Within each row, the first two columns display the source image pairs, while the remaining columns show the fused results of different comparison methods, with our fusion result placed in the rightmost column.

Benefiting from the proposed 1D signal-level decomposer and the integral-driven loss $L_{\mathrm{dec}}$, our method yields fused images that retain both the global structures and modality-specific details.
On the MIF datasets, our results preserve sharp boundaries, bone regions and texture from MRI and CT, while maintaining the functional patterns of PET and SPECT without color distortion.
On the VIF datasets, our fused images simultaneously preserve the fine textures and contrast of the visible modality and the salient thermal targets from infrared, leading to clearer edges, more stable illumination than competing methods.
These visual results are consistent with our signal-level loss which provides a clearer optimization objective with the dual pretext tasks, jointly enabling more accurate feature decomposition and, consequently, more advanced fusion results across all six datasets.

\begin{table}[t]
\centering
\caption{AP for object detection on source images and fused images from different methods. mAP is the mean of the six category APs.}
\label{tab:det_ap}
\resizebox{1\linewidth}{!}{
\begin{tabular}{c|cccccc|c}
\hline
Method & People & Car & Bus & Motorcycle & Lamp & Truck & mAP \\
\hline
IR            & 0.512 & 0.581 & 0.447 & 0.468 & 0.403 & 0.452 & 0.477 \\
VIS           & 0.563 & 0.638 & 0.483 & 0.521 & 0.436 & 0.494 & 0.522 \\
CDDFuse       & 0.662 & 0.726 & 0.618 & 0.634 & 0.573 & 0.602 & 0.636 \\
LRRNet        & 0.641 & 0.707 & 0.597 & 0.612 & 0.551 & 0.582 & 0.615 \\
EMMA          & 0.651 & 0.719 & 0.612 & 0.631 & 0.562 & 0.593 & 0.628 \\
TC\mbox{-}MoA & 0.642 & 0.715 & 0.606 & 0.623 & 0.559 & 0.601 & 0.624 \\
Text\mbox{-}DiFuse & 0.598 & 0.689 & 0.542 & 0.588 & 0.521 & 0.552 & 0.582 \\
C2RF          & 0.611 & 0.701 & 0.553 & 0.599 & 0.515 & 0.561 & 0.590 \\
SigFusion & 0.688 & \textbf{0.763} & 0.653 & 0.649 & \underline{0.579} & 0.622 & 0.655 \\
DCEvo         & \underline{0.689} & 0.735 & \textbf{0.661} & 0.653 & \textbf{0.583} & 0.612 & 0.655 \\
SAGE          & 0.629 & 0.703 & 0.571 & 0.609 & 0.542 & 0.571 & 0.604 \\
TD\mbox{-}Fusion & 0.661 & 0.731 & 0.632 & 0.643 & \textbf{0.583} & 0.611 & 0.643 \\
Omni\mbox{-}Fuse & 0.649 & 0.721 & 0.601 & 0.621 & 0.571 & 0.603 & 0.628 \\
\hline
Ours          & \textbf{0.691} & \underline{0.737} & \underline{0.658} & \textbf{0.651} & 0.577 & \textbf{0.623} & \textbf{0.656} \\
\hline
\end{tabular}}
\end{table}

\begin{table}[t]
\centering
\caption{Quantitative comparison for medical image segmentation. }
\label{tab:seg_neworder}
\resizebox{\linewidth}{!}{
\begin{tabular}{c|ccccccc}
\hline
\textbf{Metrics} & T1 & Flair & CDDFuse & LRRNet & EMMA & TC\mbox{-}MoA & Text\mbox{-}Diffuse \\
\hline
IoU  & 0.615 & 0.657 & 0.762 & 0.731 & 0.743 & 0.689 & 0.685 \\
Dice & 0.607 & 0.727 & \underline{0.850} & 0.836 & 0.841 & 0.829 & 0.751 \\
\hline
\textbf{Metrics} & CCF & SigFusion & BSAFusion & Mask\mbox{-}Difuser & MTG\mbox{-}Fusion & C2RF & Ours \\
\hline
IoU  & 0.661 & 0.735 & 0.745 & 0.723 & \underline{0.775} & 0.772 & \textbf{0.790} \\
Dice & 0.752 & 0.770 & 0.832 & 0.829 & 0.846 & 0.848 & \textbf{0.856} \\
\hline
\end{tabular}}
\end{table}

\begin{figure*}[t]
    \centering
    \includegraphics[width=\linewidth]{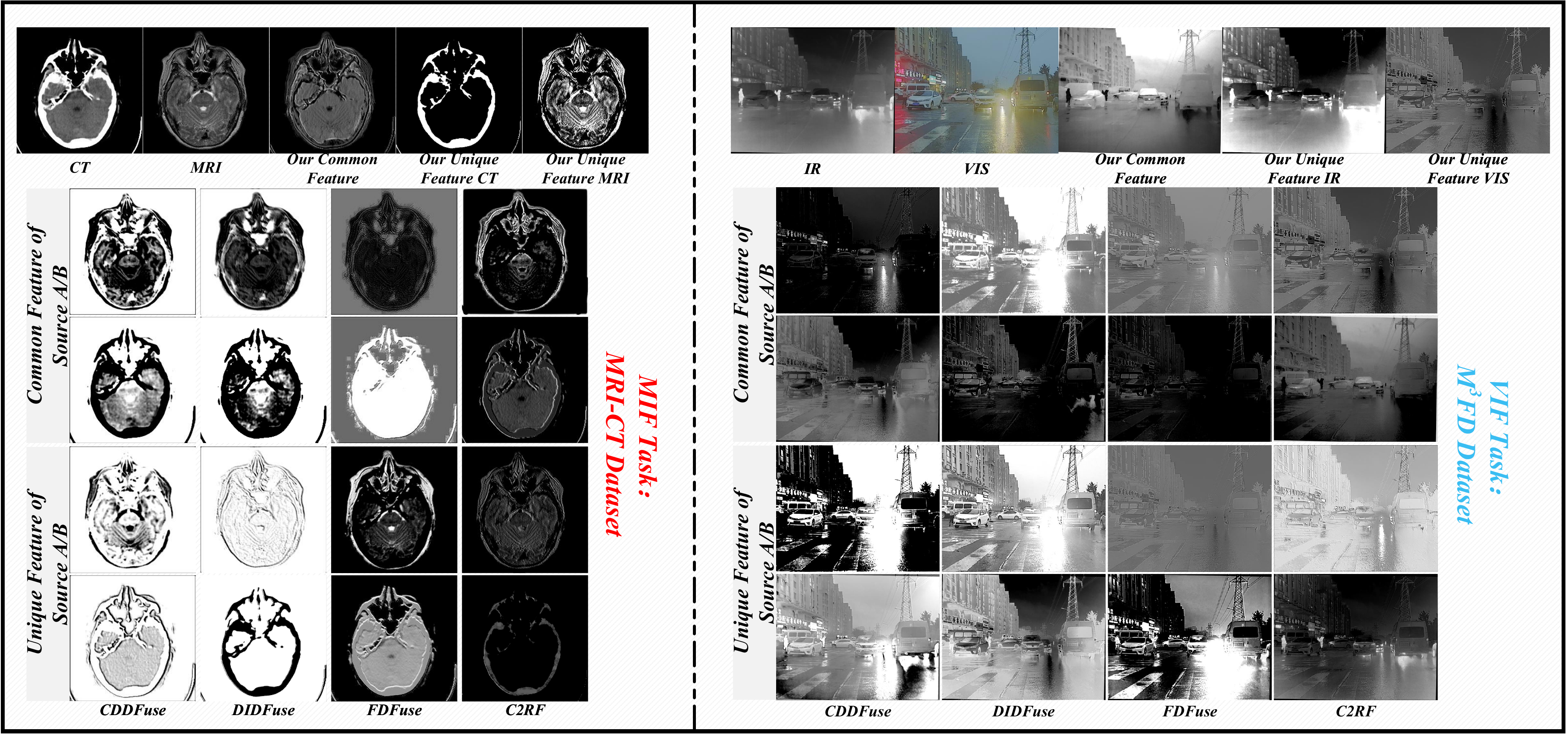}
    \caption{Visualization of common and unique feature maps produced by different decomposition strategies. For each source pair, we display the estimated common features and the modality-unique features. Compared with existing methods that generate two inconsistent common features, our signal-level decomposer yields a single common map and better-separated unique maps.}
    \label{fig:feature}
\end{figure*}
\subsection{Downstream Task Evaluation}

\textbf{Object Detection for VIF Task.}\label{de}
To evaluate whether the fused images can better support downstream visual understanding, we conduct object detection on the VIF task. Specifically, fused images generated by different VIF methods are fed into the same pre-trained YOLOv12 detector~\cite{tian2025yolov12}, without modifying the detector or fine-tuning it on any specific fusion result. This setting ensures that the comparison mainly reflects the influence of different fusion outputs on detection performance.

As shown in Fig.~\ref{fig:downstream}(left), the detection results produced from our fused images are generally more complete and accurate than those from other fusion methods. In challenging scenes, competing methods may miss small targets, generate inaccurate bounding boxes, or confuse objects with weak contrast. In contrast, our method better preserves complementary infrared and visible cues, which helps the detector localize objects more reliably. Table~\ref{tab:det_ap} reports the per-class AP~\cite{lin2014microsoft} and mAP results. Compared with other VIF methods, our method achieves the best overall mAP and competitive AP on most object categories, demonstrating that the proposed fusion strategy can provide more task-favorable inputs for downstream detection.

\textbf{Medical Image Segmentation for MIF Task.}\label{seg}
We further evaluate the downstream utility of our method on medical image segmentation. In this experiment, the fused images generated by different MIF methods are used as inputs to UniverSeg~\cite{butoi2023universeg}, and the segmentation results are compared with the ground-truth masks. This evaluation examines whether the fused images can preserve lesion-related structures and boundary information that are useful for medical analysis.

As shown in Fig.~\ref{fig:downstream}(right), segmentation results based on our fused images are closer to the ground truth, especially around lesion boundaries and structurally complex regions. Other fusion methods may produce incomplete lesion regions or less accurate boundaries due to insufficient preservation of cross-modal complementary information. In contrast, our method better maintains common anatomical structures while retaining modality-specific details, which provides more discriminative cues for segmentation. Table~\ref{tab:seg_neworder} reports the IoU~\cite{ZOU2004178} and Dice~\cite{milletari2016v} results. Our method obtains the best performance among the compared fusion methods, confirming that the proposed decomposition strategy can improve not only visual fusion quality but also downstream medical segmentation performance.

\subsection{Visualization of Common and Unique Features}
To better understand our feature decomposition mechanism, we compare the estimated common and unique features of our method with those of several representative decomposition-based baselines.
For each source pair $(I_A, I_B)$, we visualize the common feature maps and the corresponding modality-unique feature maps.

As shown in Fig.~\ref{fig:feature}, existing image-level decomposers typically produce two common features for each modality. 
In practice, these two features are not strictly shared as they often contain modality-specific patterns, which contradicts the assumption that there should be a single common component underlying both modalities.
In contrast, our signal-level decomposer explicitly models a single common feature map $C_{AB}$ together with two unique feature maps for $I_A$ and $I_B$, leading to a more reasonable and identifiable decomposition form.

The visualizations further reveal that, for our method, the common feature mainly captures shared semantic content and global structures, while the unique feature concentrate on modality-specific cues.
These observations align with our signal-level formulation and support that the proposed decomposer separates common and modality-specific information in a more structured and disentangled way.

\subsection{t-SNE Visualization of Frequency and Decomposed Features}

\begin{figure}[t]
    \centering
    \includegraphics[width=0.75\linewidth]{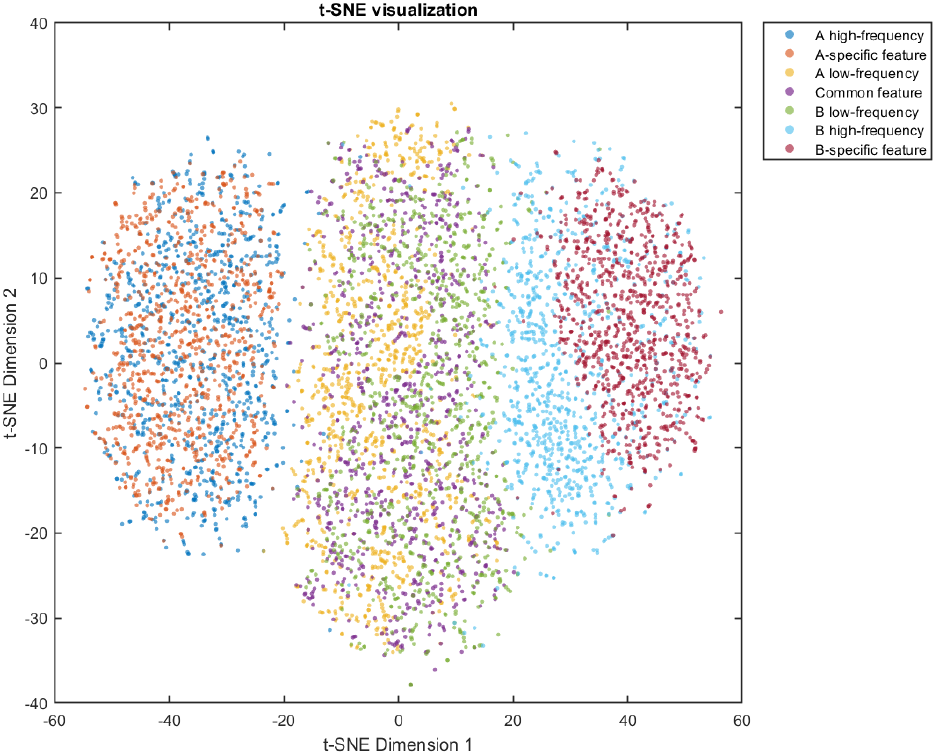}
    \caption{
    t-SNE visualization of low-/high-frequency components and learned decomposed features. The common features are close to the low-frequency components of both modalities, while the unique features are closer to their corresponding high-frequency components, supporting the design of extracting common information from low frequencies and unique information from high frequencies.
    }
    \label{fig:tsne}
\end{figure}

To further analyze the relationship between frequency components and the learned decomposed features, we visualize their feature distributions using t-SNE. Specifically, we project the low-frequency components, high-frequency components, learned common features, and modality-specific unique features into a two-dimensional space. As shown in Fig.~\ref{fig:tsne}, the learned common features are mainly distributed around the low-frequency components of both modalities, indicating that low-frequency signals contain more modality-shared structural information. In contrast, the learned unique features are closer to the corresponding high-frequency components and form more modality-specific distributions. This observation is consistent with our design motivation that common information can be better extracted from low-frequency components, while modality-specific details are more related to high-frequency components.

These results provide intuitive evidence for the proposed decomposition strategy. By exploiting the complementary roles of low- and high-frequency components, our method can more effectively separate common structures and modality-specific details.

\begin{figure*}[t]
    \centering
    \includegraphics[width=0.95\textwidth]{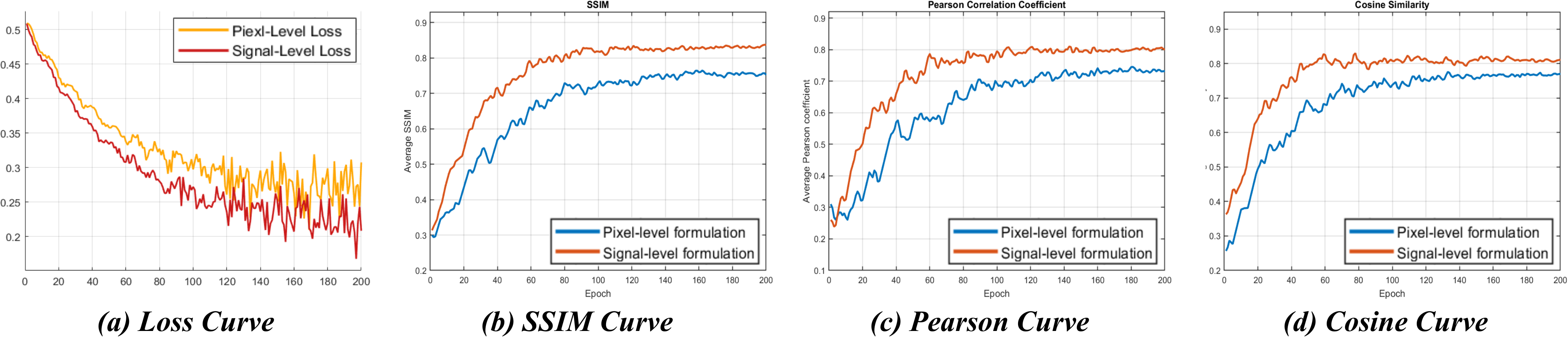}
    \caption{
    Visualization of training curves. We compare the proposed signal-level objective with a conventional 2D image-domain mixed-loss strategy in terms of training loss and feature similarity, including SSIM, Pearson correlation, and cosine similarity. The results show that our signal-level objective converges faster and achieves a lower final loss, while maintaining higher and more stable similarity between the learned common features and source signals.
    }
    \label{fig:training_curves}
\end{figure*}

\subsection{Visualization of Training Curves}
To analyze the optimization behavior of the proposed signal-level formulation, we compare it with a conventional 2D image-domain mixed-loss strategy. As shown in Fig.~\ref{fig:training_curves}, our signal-level objective converges faster and reaches a lower final loss, indicating a clearer and more stable optimization target.

We further visualize the similarity curves during training, including SSIM, Pearson correlation, and cosine similarity. Compared with the 2D mixed-loss strategy, our method shows higher and more stable similarity between the learned common features and source signals. These results provide intuitive evidence that the proposed signal-level objective can better guide common feature learning.
\begin{table}
\centering
\caption{The refinement effects results of our paradigm. Original vs.\ upgraded models (* denotes our paradigm inserted). }
\label{tab:broader_plug_in}
\resizebox{1\textwidth}{!}{%
\begin{tabular}{l|ccccccc}
\hline
Method & $Q_{MI}\!\uparrow$ & $Q_{NICE}\!\uparrow$ & $Q_{P}\!\uparrow$ & $Q_{CB}\!\uparrow$ & $MI\!\uparrow$ & $VIF_p\!\uparrow$ & $Q_{Y}\!\uparrow$ \\
\hline
DIDFuse    & 0.6130 & 0.8083 & 0.3570 & 0.4830 & 3.2370 & 0.3610 & 0.9020 \\
DIDFuse*   & 0.7466 & 0.8196 & 0.4198 & 0.5299 & 3.9038 & 0.4596 & 0.9101 \\
\textbf{Gain} & \color{red}21.8\%\,$\uparrow$ & \color{red}1.4\%\,$\uparrow$  & \color{red}17.6\%\,$\uparrow$ & \color{red}9.7\%\,$\uparrow$  & \color{red}20.6\%\,$\uparrow$ & \color{red}27.3\%\,$\uparrow$ & \color{red}0.9\%\,$\uparrow$  \\
\hline
CDDFuse    & 0.6610 & 0.8107 & 0.4010 & 0.6940 & 3.9820 & 0.3990 & 0.9380 \\
CDDFuse*   & 0.7582 & 0.8277 & 0.4523 & 0.7051 & 4.3563 & 0.4736 & 0.9446 \\
\textbf{Gain} & \color{red}14.7\%\,$\uparrow$ & \color{red}2.1\%\,$\uparrow$  & \color{red}12.8\%\,$\uparrow$ & \color{red}1.6\%\,$\uparrow$  & \color{red}9.4\%\,$\uparrow$  & \color{red}18.7\%\,$\uparrow$ & \color{red}0.7\%\,$\uparrow$  \\
\hline
FDFuse     & 0.6420 & 0.8086 & 0.3890 & 0.6860 & 3.9050 & 0.3920 & 0.9330 \\
FDFuse*    & 0.8044 & 0.8215 & 0.4820 & 0.7320 & 4.5532 & 0.5088 & 0.9433 \\
\textbf{Gain} & \color{red}25.3\%\,$\uparrow$ & \color{red}1.6\%\,$\uparrow$  & \color{red}23.9\%\,$\uparrow$ & \color{red}6.7\%\,$\uparrow$  & \color{red}16.6\%\,$\uparrow$ & \color{red}29.8\%\,$\uparrow$ & \color{red}1.1\%\,$\uparrow$  \\
\hline
C2RF       & 0.6360 & 0.8071 & 0.3660 & 0.6740 & 3.8510 & 0.3730 & 0.9290 \\
C2RF*      & 0.8243 & 0.8224 & 0.4663 & 0.7535 & 4.8022 & 0.4566 & 0.9383 \\

\textbf{Gain} & \color{red}29.6\%\,$\uparrow$ & \color{red}1.9\%\,$\uparrow$  & \color{red}27.4\%\,$\uparrow$ & \color{red}11.8\%\,$\uparrow$ & \color{red}24.7\%\,$\uparrow$ & \color{red}22.4\%\,$\uparrow$ & \color{red}1.0\%\,$\uparrow$  \\

\hline
\end{tabular}%
}
\end{table}

\section{Additional Results for Broader Impact}
\label{sec:appendix_broader}

In this section, we provide the detailed quantitative results corresponding to the broader impact analysis in the main paper. We mainly consider two aspects: refining existing decomposition-based MMIF paradigms and extending the proposed signal-level decomposition paradigm to other image fusion tasks.

\subsection{Refining Decomposition-based MMIF Paradigms}
\label{sec:appendix_refine_mmif}

To further validate the compatibility of the proposed signal-level decomposition paradigm, we insert it into several representative decomposition-based MMIF models. For a fair comparison, we keep the original training strategies, network settings, and evaluation protocols of these models unchanged, and only replace or enhance their feature decomposition process with our signal-level decomposition paradigm. The enhanced variants are denoted by ``$^\ast$''.

Table~\ref{tab:broader_plug_in} reports the quantitative results. Compared with the original models, the enhanced variants achieve consistent improvements on most evaluation metrics. These results indicate that the proposed paradigm is not limited to our specific network architecture. Instead, by providing a clearer objective for separating common and unique features, it can serve as a generally effective refinement for existing decomposition-based MMIF methods.

\begin{table*}[t]
\centering
\small
\caption{Quantitative comparison on the MFIF task against 9 competing methods using 6 evaluation metrics on the LYTRO and MFFW datasets. Although MFIF may provide fused ground truth for evaluation, the decomposition of common and unique features still lacks direct supervision. Our method remains applicable in this setting and achieves strong overall performance.}
\label{tab:mfif}
\resizebox{\textwidth}{!}{%
\begin{tabular}{l l cccccc | cccccc}
\toprule
\multirow{2}{*}{Method} & \multirow{2}{*}{Pub/Year}
& \multicolumn{6}{c|}{Dataset: LYTRO}
& \multicolumn{6}{c}{Dataset: MFFW} \\
\cmidrule(lr){3-8} \cmidrule(lr){9-14}
& 
& $Q_G \uparrow$ & $Q_M \uparrow$ & $Q_P \uparrow$ & $MI \uparrow$ & $SD \uparrow$ & $VIFF \uparrow$
& $Q_G \uparrow$ & $Q_M \uparrow$ & $Q_P \uparrow$ & $MI \uparrow$ & $SD \uparrow$ & $VIFF \uparrow$ \\
\midrule
CUNet      & TPAMI 20  & 0.526 & 0.553 & 0.696 & 5.441 & \underline{58.70} & 1.022 & 0.482 & 0.455 & 0.552 & 4.593 & \underline{56.33} & 0.847 \\
U2Fusion   & TPAMI 20  & 0.580 & 0.480 & 0.742 & 5.677 & 58.37 & 1.086 & 0.537 & 0.405 & 0.611 & 4.876 & 55.26 & 0.825 \\
DeFusion   & ECCV 22   & 0.455 & 0.325 & 0.660 & 5.984 & 54.39 & 1.028 & 0.418 & 0.296 & 0.518 & 5.137 & 51.55 & 0.876 \\
DIFNet     & CVPR 22   & 0.437 & 0.325 & 0.688 & 5.774 & 49.67 & 1.032 & 0.422 & 0.309 & 0.577 & 4.867 & 46.66 & 0.890 \\
FusionDiff & ESWA 23   & 0.629 & 0.821 & 0.783 & 6.554 & 56.13 & 1.188 & 0.545 & 0.602 & 0.659 & 5.334 & 53.27 & 0.993 \\
MGDN       & ACMMM 23  & \textbf{0.662} & 0.901 & 0.810 & 6.655 & 56.88 & 1.226 & \underline{0.606} & 0.650 & 0.663 & \underline{5.558} & 54.47 & 1.020 \\
ZMFF       & INFFUS 23 & 0.631 & 0.600 & 0.785 & 6.235 & 57.06 & 1.175 & 0.552 & 0.487 & 0.635 & 5.092 & 54.38 & 0.990 \\
DeepM2CDL  & TPAMI 24  & 0.639 & \underline{1.004} & \underline{0.810} & 6.441 & 58.05 & 1.267 & 0.582 & \underline{0.805} & \underline{0.679} & 5.372 & 56.01 & \underline{1.064} \\
FILM       & ICML 24   & 0.619 & 0.567 & 0.782 & \textbf{6.758} & \textbf{59.15} & \underline{1.283} & 0.498 & 0.436 & 0.544 & 5.254 & \textbf{57.10} & 0.919 \\
Ours       & -         & \underline{0.659} & \textbf{1.031} & \textbf{0.811} & \underline{6.733} & 58.66 & \textbf{1.292} & \textbf{0.611} & \textbf{0.994} & \textbf{0.682} & \textbf{6.002} & 56.18 & \textbf{1.112} \\
\bottomrule
\end{tabular}%
}
\end{table*}

\subsection{Extension to Multi-focus Image Fusion}
\label{sec:appendix_mfif}

We further examine whether the proposed signal-level decomposition paradigm can be transferred to other image fusion tasks beyond MMIF. In this section, we select multi-focus image fusion as a representative task. Different from MMIF, multi-focus image fusion aims to integrate multiple images of the same scene with different focus regions. Although the imaging setting is different, this task also requires the model to preserve common scene structures while extracting source-specific focused details. Therefore, multi-focus image fusion provides a suitable testbed for evaluating the broader applicability of our decomposition paradigm.

Table~\ref{tab:mfif} reports the quantitative results. The proposed paradigm achieves competitive or superior performance across multiple metrics. These results are consistent with the qualitative observations in the main paper, where our method produces clearer focus boundaries and more faithful structural details. This further demonstrates that the signal-level decomposition paradigm can be naturally extended to other image fusion tasks and has potential value for more general fusion problems.

\begin{table*}[t]
\centering
\caption{Quantitative results of multiple ablation experiments for VIF and MIF tasks on six datasets.}
\label{tab:ablation_six_datasets}
\resizebox{\textwidth}{!}{%
\begin{tabular}{l|ccccccc|ccccccc}
\hline
\multicolumn{15}{c}{\textbf{I. Effect of Signal-Level Decomposer}}\\
\hline
 & \multicolumn{7}{c|}{VIF task} & \multicolumn{7}{c}{MIF task} \\
Configuration settings 
& $Q_{MI}$↑ & $Q_{NICE}$↑ & $Q_{P}$↑ & $Q_{CB}$↑ & $MI$↑ & $VIF_p$↑ & $Q_{Y}$↑
& $Q_{MI}$↑ & $Q_{NICE}$↑ & $Q_{P}$↑ & $Q_{CB}$↑ & $MI$↑ & $VIF_p$↑ & $Q_{Y}$↑ \\
\hline
\multicolumn{1}{c|}{} & \multicolumn{7}{c|}{\emph{VIF: M$^3$FD}} & \multicolumn{7}{c}{\emph{MIF: MRI-CT}}\\
w/o signal-level decomposer        
& 0.6282 & 0.8181 & 0.5350 & 0.4716 & 3.8631 & 0.4391 & 0.8729
& 0.9012 & 0.8071 & 0.3627 & 0.6715 & 3.6129 & 0.3689 & 0.9127 \\
w/o DWT                            
& 0.6425 & 0.8192 & 0.5551 & 0.4860 & 3.9812 & 0.4534 & 0.8847
& 0.9185 & 0.8090 & 0.3825 & 0.6880 & 3.8310 & 0.3897 & 0.9312 \\
w/ IDWT $\leftarrow$ learnable reconstructor  
& \underline{0.6531} & \underline{0.8198} & \underline{0.5631} & \underline{0.4925} & \underline{4.0200} & \underline{0.4594} & \underline{0.8911}
& \underline{0.9253} & \underline{0.8096} & \underline{0.3928} & \underline{0.6905} & \underline{3.9025} & \underline{0.3954} & \underline{0.9361} \\
w/ signal-level $\leftarrow$ image-level decomposer 
& 0.6380 & 0.8185 & 0.5460 & 0.4793 & 3.9022 & 0.4468 & 0.8781
& 0.9108 & 0.8086 & 0.3755 & 0.6810 & 3.7420 & 0.3798 & 0.9275 \\
Default setups (ours)              
& \textbf{0.6608} & \textbf{0.8209} & \textbf{0.5735} & \textbf{0.4998} & \textbf{4.1050} & \textbf{0.4694} & \textbf{0.8988}
& \textbf{0.9369} & \textbf{0.8108} & \textbf{0.4034} & \textbf{0.7008} & \textbf{4.0184} & \textbf{0.4034} & \textbf{0.9439} \\
\hline
\multicolumn{1}{c|}{} & \multicolumn{7}{c|}{\emph{VIF: MSRS}} & \multicolumn{7}{c}{\emph{MIF: MRI-PET}}\\
w/o signal-level decomposer        
& 0.8154 & 0.8230 & 0.5112 & 0.4667 & 4.8120 & 0.4932 & 0.8791 
& 0.8022 & 0.8053 & 0.4613 & 0.4442 & 2.8931 & 0.4351 & 0.5620 \\
w/o DWT                            
& 0.8260 & 0.8268 & 0.5304 & 0.5551 & 5.0921 & 0.5085 & 0.8897
& 0.8191 & 0.8061 & 0.4780 & 0.4557 & 2.9874 & 0.4467 & 0.5734 \\
w/ IDWT $\leftarrow$ learnable reconstructor  
& \underline{0.8335} & \underline{0.8279} & \underline{0.5411} & \underline{0.5645} & \underline{5.1982} & \underline{0.5166} & \underline{0.8935}
& \underline{0.8318} & \underline{0.8070} & \underline{0.4876} & \underline{0.4621} & \underline{3.0315} & \underline{0.4520} & \underline{0.5826} \\
w/ signal-level $\leftarrow$ image-level decomposer 
& 0.8213 & 0.8250 & 0.5283 & 0.5512 & 5.0455 & 0.5033 & 0.8860
& 0.8244 & 0.8066 & 0.4762 & 0.4520 & 2.9634 & 0.4447 & 0.5692 \\
Default setups (ours)              
& \textbf{0.8405} & \textbf{0.8290} & \textbf{0.5483} & \textbf{0.5798} & \textbf{5.4547} & \textbf{0.5253} & \textbf{0.8936}
& \textbf{0.8843} & \textbf{0.8074} & \textbf{0.4967} & \textbf{0.4740} & \textbf{3.0547} & \textbf{0.4577} & \textbf{0.5808} \\
\hline
\multicolumn{1}{c|}{} & \multicolumn{7}{c|}{\emph{VIF: TNO}} & \multicolumn{7}{c}{\emph{MIF: MRI-SPECT}}\\
w/o signal-level decomposer        
& 0.5410 & 0.8105 & 0.4023 & 0.5115 & 3.7412 & 0.4544 & 0.8322
& 0.9621 & 0.8072 & 0.5260 & 0.6521 & 2.9435 & 0.4410 & 0.8998 \\
w/o DWT                            
& 0.5567 & 0.8116 & 0.4195 & 0.5237 & 3.7952 & 0.4658 & 0.8390
& 0.9699 & 0.8079 & 0.5361 & 0.6615 & 2.9857 & 0.4478 & 0.9047 \\
w/ IDWT $\leftarrow$ learnable reconstructor  
& \underline{0.5630} & \underline{0.8121} & \underline{0.4264} & \underline{0.5298} & \underline{3.8434} & \underline{0.4722} & \underline{0.8424}
& \underline{0.9755} & \underline{0.8080} & \underline{0.5445} & \underline{0.6659} & \underline{3.0213} & \underline{0.4511} & \underline{0.9101} \\
w/ signal-level $\leftarrow$ image-level decomposer 
& 0.5489 & 0.8113 & 0.4152 & 0.5189 & 3.7680 & 0.4607 & 0.8361
& 0.9682 & 0.8075 & 0.5324 & 0.6582 & 2.9880 & 0.4482 & 0.9036 \\
Default setups (ours)              
& \textbf{0.5711} & \textbf{0.8128} & \textbf{0.4318} & \textbf{0.5346} & \textbf{3.8015} & \textbf{0.4768} & \textbf{0.8450}
& \textbf{0.9778} & \textbf{0.8083} & \textbf{0.5507} & \textbf{0.6697} & \textbf{3.0655} & \textbf{0.4520} & \textbf{0.9149} \\
\hline\hline
\multicolumn{15}{c}{\textbf{II. Effect of Dual Pretext Tasks}}\\
\hline
 & \multicolumn{7}{c|}{VIF task} & \multicolumn{7}{c}{MIF task} \\
Configuration settings 
& $Q_{MI}$↑ & $Q_{NICE}$↑ & $Q_{P}$↑ & $Q_{CB}$↑ & $MI$↑ & $VIF_p$↑ & $Q_{Y}$↑
& $Q_{MI}$↑ & $Q_{NICE}$↑ & $Q_{P}$↑ & $Q_{CB}$↑ & $MI$↑ & $VIF_p$↑ & $Q_{Y}$↑ \\
\hline
\multicolumn{1}{c|}{} & \multicolumn{7}{c|}{\emph{VIF: M$^3$FD}} & \multicolumn{7}{c}{\emph{MIF: MRI-CT}}\\
w/o signal-level pretext task      
& 0.6315 & 0.8188 & 0.5420 & \underline{0.4868} & 3.9033 & 0.4512 & 0.8799
& 0.9144 & 0.8093 & 0.3786 & \underline{0.6869} & 3.7562 & 0.3831 & 0.9293 \\
w/o image-level pretext task       
& \underline{0.6495} & \underline{0.8195} & \underline{0.5525} & 0.4820 & \underline{3.9800} & \underline{0.4553} & \underline{0.8850}
& \underline{0.9228} & \textbf{0.8294} & \underline{0.3866} & 0.6765 & \underline{3.8041} & \underline{0.3912} & \underline{0.9335} \\
w/ $L_{\mathrm{dec}}{\leftarrow}$ $L_{1}$ 
& 0.6463 & 0.8195 & 0.5482 & 0.4817 & 3.9521 & 0.4520 & 0.8832
& 0.9202 & 0.8091 & 0.3814 & 0.6851 & 3.7820 & 0.3886 & 0.9318 \\
Default setups (ours)              
& \textbf{0.6608} & \textbf{0.8209} & \textbf{0.5735} & \textbf{0.4998} & \textbf{4.1050} & \textbf{0.4694} & \textbf{0.8988}
& \textbf{0.9369} & \underline{0.8108} & \textbf{0.4034} & \textbf{0.7008} & \textbf{4.0184} & \textbf{0.4034} & \textbf{0.9439} \\
\hline
\multicolumn{1}{c|}{} & \multicolumn{7}{c|}{\emph{VIF: MSRS}} & \multicolumn{7}{c}{\emph{MIF: MRI-PET}}\\
w/o signal-level pretext task      
& 0.8128 & 0.8220 & 0.5287 & 0.5652 & 5.1053 & \underline{0.5188} & 0.8865
& 0.8230 & 0.8062 & \textbf{0.5025} & \underline{0.4604} & 2.9634 & 0.4461 & 0.5742 \\
w/o image-level pretext task       
& \textbf{0.8467} & \underline{0.8265} & \underline{0.5361} & \underline{0.5661} & \underline{5.1980} & 0.5134 & \underline{0.8922}
& \underline{0.8322} & \underline{0.8070} & 0.4860 & 0.4583 & \underline{3.0258} & \underline{0.4523} & \textbf{0.5831} \\
w/ $L_{\mathrm{dec}}{\leftarrow}$ $L_{1}$ 
& 0.8223 & 0.8241 & 0.5338 & 0.5594 & 5.1647 & 0.5110 & 0.8895
& 0.8285 & 0.8067 & 0.4822 & 0.4562 & 3.0043 & 0.4495 & 0.5796 \\
Default setups (ours)              
& \underline{0.8405} & \textbf{0.8290} & \textbf{0.5483} & \textbf{0.5798} & \textbf{5.4547} & \textbf{0.5253} & \textbf{0.8936}
& \textbf{0.8843} & \textbf{0.8074} & \underline{0.4967} & \textbf{0.4740} & \textbf{3.0547} & \textbf{0.4577} & \underline{0.5808} \\
\hline
\multicolumn{1}{c|}{} & \multicolumn{7}{c|}{\emph{VIF: TNO}} & \multicolumn{7}{c}{\emph{MIF: MRI-SPECT}}\\
w/o signal-level pretext task      
& 0.5468 & 0.8105 & 0.4140 & \underline{0.5235} & 3.7845 & 0.4620 & 0.8367
& \underline{0.9759} & 0.8075 & 0.5422 & \underline{0.6631} & 3.0176 & 0.4505 & 0.9085 \\
w/o image-level pretext task       
& 0.5590 & \underline{0.8118} & \underline{0.4256} & 0.5212 & \underline{3.8210} & \underline{0.4701} & \underline{0.8412}
& 0.9758 & \underline{0.8080} & \underline{0.5480} & 0.6610 & \underline{3.0435} & \textbf{0.4526} & \underline{0.9120} \\
w/ $L_{\mathrm{dec}}{\leftarrow}$ $L_{1}$ 
& \textbf{0.5727} & 0.8112 & 0.4203 & 0.5194 & 3.7958 & 0.4672 & 0.8389
& 0.9747 & 0.8078 & 0.5451 & 0.6598 & 3.0299 & 0.4516 & 0.9103 \\
Default setups (ours)              
& \underline{0.5711} & \textbf{0.8128} & \textbf{0.4318} & \textbf{0.5346} & \textbf{3.8015} & \textbf{0.4768} & \textbf{0.8450}
& \textbf{0.9778} & \textbf{0.8083} & \textbf{0.5507} & \textbf{0.6697} & \textbf{3.0655} & \underline{0.4520} & \textbf{0.9149} \\
\hline
\end{tabular}%
}
\end{table*}

% ================== NEXT MAIN SECTION ==================
\section{Extended Ablation Study}
\label{sec:more abl}
\begin{figure}[t]
    \centering
    \includegraphics[width=\linewidth]{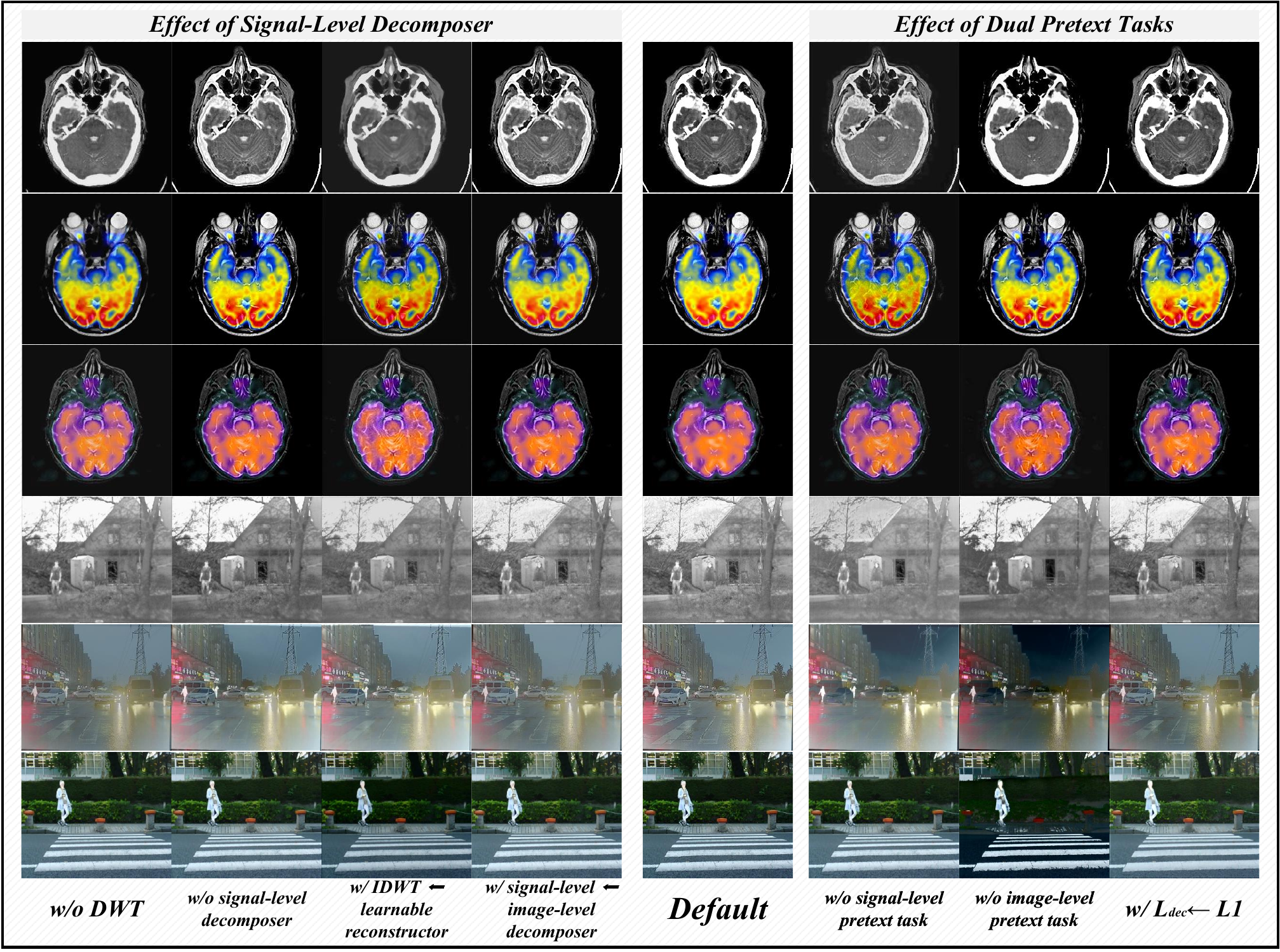}
    \caption{Visualization of ablation results for different configurations on representative VIF and MIF cases.}
    \label{fig:Ablation}
\end{figure}

To comprehensively verify the effectiveness of each key component, we conduct an extended ablation study on all six datasets for both VIF and MIF tasks. The quantitative results of all variants are summarized in Table \ref{tab:ablation_six_datasets}, and representative qualitative comparisons for typical cases are visualized in Fig. \ref{fig:Ablation}, complementing the main-paper results.

\begin{figure*}
    \centering
    \includegraphics[width=1\linewidth]{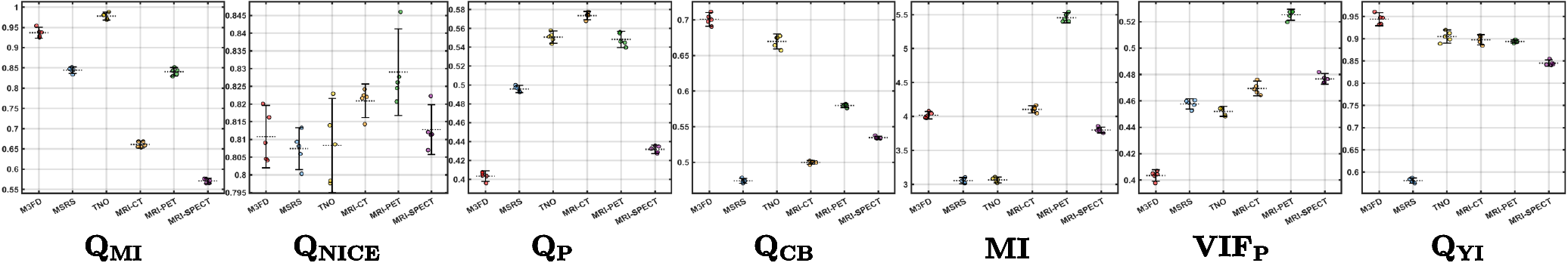}
    \caption{Reproducibility verification under different random seeds. For each metric, markers denote independent runs with different initializations, and their tight clustering with small standard deviations indicates that our method is robust to random initialization.}
    \label{fig:placeholder}
\end{figure*}

\begin{figure*}
    \centering
    \includegraphics[width=1\linewidth]{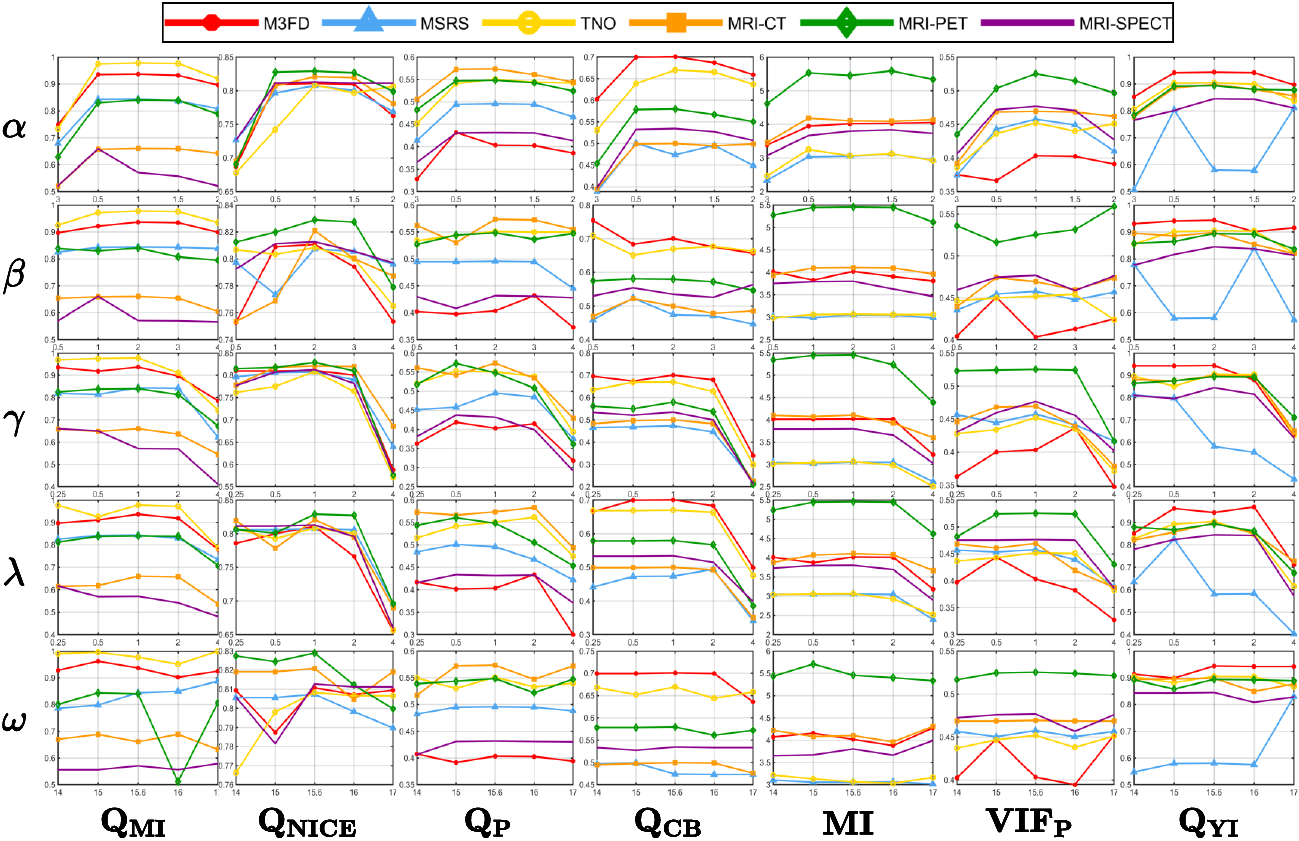}
    \caption{Sensitivity to key hyperparameters.}
    \label{fig:Sensitivity}
\end{figure*}

\paragraph{Effect of Signal-Level Decomposer.}
Our signal-level decomposer transfers the decomposition problem into 1D signal-level, effectively disentangling image-level coupling and enabling more accurate separation of features. To assess its impact, we conduct the following ablations:
(1) removing the signal-level decomposer;
(2) removing frequency decomposition;
(3) replacing IDWT with a learnable reconstructor;
(4) substituting an image-level decomposer;
(5) default setup.
The corresponding quantitative results are given in the upper part of Table~\ref{tab:ablation_six_datasets}. Across all six datasets and seven metrics, the default setup consistently achieves the advanced performance, while removing or weakening any part of the signal-level decomposer leads to clear drops, which further demonstrates the effectiveness of our paradigm.
In Fig.~\ref{fig:Ablation} (left), we visualize typical VIF and MIF examples under different configurations: the variants without a complete signal-level decomposer tend to produce blurred structures, noisy background or loss of weak details, whereas the default setup preserves sharper organ boundaries, clearer edges and richer complementary cues from both modalities.
\paragraph{Effect of Dual Pretext Tasks.}
To evaluate the dual pretext tasks at both levels, we perform the following ablations:
(1) removing the signal-level pretext task;
(2) removing the image-level pretext task;
(3) replacing the proposed $L_{\mathrm{dec}}$ with vanilla $L_1$ only;
(4) default configuration.
The results are reported in the lower part of Table~\ref{tab:ablation_six_datasets}. Removing either pretext task consistently degrades performance on all datasets, and using a naive $L_1$ loss also weakens the results, highlighting the necessity to provide the decomposer a clear and properly designed supervision. These tendencies are also clearly reflected in Fig.~\ref{fig:Ablation} (right), where the default configuration yields the most faithful results.

\paragraph{Sensitivity to Key Hyperparameters.}
\label{sensitivity}
We further analyze the sensitivity of key hyperparameters and design choices. The corresponding curves and additional ablation plots are provided in Fig. \ref{fig:Sensitivity} of the Appendix.

% ================== NEXT MAIN SECTION ==================
\section{Reproducibility Verification}
We further conduct robustness checks with respect to random initialization. 
Keeping all architectural settings unchanged, we rerun our model with multiple random seeds. 
As shown in Fig. \ref{fig:placeholder}, the results under different seeds are tightly clustered, and the quantitative metrics exhibit small standard deviations, indicating that our method is stable.

% ================== NEXT MAIN SECTION ==================
% \section{Statistical Significance Analysis}
% We conduct paired $t$-tests between our method and each baseline on all benchmarks.
% For each metric and dataset, we report the $p$-values and indicate statistically significant improvements (e.g., $p < 0.05$).
% The results confirm that the gains achieved by our approach are not due to random fluctuations but are statistically reliable across test sets.

% ================== NEXT MAIN SECTION ==================
\section{Model Complexity and Overhead Analysis}
We analyze the computational overhead, including the number of parameters, GFLOPs, and inference time as shown in Table \ref{tab:complexity}.
Compared with representative baselines, our model remains computationally competitive. Although the use of DWT introduces some additional overhead, the resulting performance gains make this cost well justified.

% ================== NEXT MAIN SECTION ==================
\section{Limitations}
While our algorithm demonstrates promising performance, it still has several limitations. First, our method currently relies on pre-registered image pairs to achieve optimal fusion results. This requirement, which is also shared by many mainstream image fusion models, may restrict its applicability in scenarios where accurate pre-registration is challenging or expensive. Second, since our method operates in the signal domain, larger images, when flattened, produce longer 1D signals and thus lead to relatively slower processing. This indicates that image size has an impact on the computational efficiency of our algorithm, which suggests a promising direction for future optimization.

\begin{table}[t]
\centering
\caption{Comparison of runtime, computational cost (GFLOPs), and model size (Params) on VIF and MIF tasks. All values are averaged per image.}
\label{tab:complexity}
\resizebox{\linewidth}{!}{%
\begin{tabular}{lccc|lccc}
\toprule
\multicolumn{4}{c|}{\textbf{VIF}} & \multicolumn{4}{c}{\textbf{MIF}} \\ 
\cmidrule(lr){1-4} \cmidrule(lr){5-8}
\textbf{Method} & \textbf{Time (s)} & \textbf{GFLOPs (G)} & \textbf{Params (M)} &
\textbf{Method} & \textbf{Time (s)} & \textbf{GFLOPs (G)} & \textbf{Params (M)} \\
\midrule
CDDFuse       & 0.006  & 116.851  & 1.186 & CDDFuse      & 0.025  & 116.851  & 1.186 \\
LRRNet       & 0.001  & 0.001    & 0.049 & LRRNet      & 0.001  & 0.001    & 0.049 \\
EMMA         & 0.058  & 8.861    & 1.516 & EMMA        & 0.009  & 8.861    & 1.516 \\
TC-MoA       & 0.543  & 61.000   & 340.580 & TC-MoA    & 0.161  & 61.000   & 340.580 \\
Text-Difuse & 23.818 & 18513 & 119.460 & Text-Difuse & 24.424 & 2742.5 & 119.460 \\
SigFusion      & 0.533  & 41.761   & 0.177 & CCF         & 31.876 & 1114.000 & 552.810 \\
DCEvo        & 0.521  & 15.000   & 2.000 & SigFusion     & 0.037  & 41.761   & 0.177 \\
SAGE         & 0.020  & 29.285   & 0.136 & BSA-Fusion   & 0.127  & 13.437   & 9.694 \\
TD-Fusion    & 0.053  & 3.886    & 0.059 & Mask-Difuser & 0.503 & 995.706  & 171.262 \\
Omni-Fuse  & 2.614  & 172.923  & 78.320 & MTG-Fusion  & 0.053  & 121.411  & 4.273 \\
C2RF         & 0.089  & 123.869  & 1.325 & C2RF        & 0.112  & 123.869  & 1.325 \\
% \rowcolor{gray!10}
\hline
Ours & 0.059 & 4.441 & 0.610 & 
Ours & 0.011 & 4.441 & 0.610 \\
\bottomrule
\end{tabular}}
\end{table}

%%%%%%%%%%%%%%%%%%%%%%%%%%%%%%%%%%%%%%%%%%%%%%%%%%%%%%%%%%%%
\iffalse
\newpage
\input{checklist.tex}
\fi

\end{document}